\documentclass[11pt]{article}
\usepackage{amsmath,amsfonts,amsmath,amssymb,amsthm}
\usepackage{comment}
\usepackage{bm}
\usepackage{graphicx,psfrag,epsf}
\usepackage{enumerate}
\usepackage{natbib}
\usepackage{algorithm}
\usepackage[noend]{algpseudocode}
\RequirePackage[colorlinks,citecolor=blue,linkcolor=blue,urlcolor=blue]{hyperref}

\usepackage{xcolor}

\usepackage{url} 
\newcommand{\blind}{0}

\usepackage{tikz}
\usepackage[capitalise]{cleveref}

\usepackage[shortlabels]{enumitem}

\newtheorem{definition}{Definition}

\newcommand{\RFS}{\text{RFS}}
\newcommand{\OLS}{\text{OLS}}

\begin{document}

\def\spacingset#1{\renewcommand{\baselinestretch}%
{#1}\small\normalsize} \spacingset{1}

\if0\blind
{
  \title{\bf Trees, Forests, Chickens, and Eggs:  When and Why to Prune Trees in a Random Forest\\}
  \author{\Large{Siyu Zhou and Lucas Mentch} \\ ~ \\ Department of Statistics \\ University of Pittsburgh}

  \maketitle
} \fi

\if1\blind
{
  \bigskip
  \bigskip
  \bigskip
  \begin{center}
    {\LARGE\bf Trees, Forests, Chickens, and Eggs:  When and Why to Prune Trees in a Random Forest}
\end{center}
  \medskip
} \fi

\bigskip
\begin{abstract}
\noindent Due to their long-standing reputation as excellent off-the-shelf predictors, random forests continue remain a go-to model of choice for applied statisticians and data scientists.  Despite their widespread use, however, until recently, little was known about their inner-workings and about which aspects of the procedure were driving their success.  Very recently, two competing hypotheses have emerged -- one based on interpolation and the other based on regularization.  This work argues in favor of the latter by utilizing the regularization framework to reexamine the decades-old question of whether individual trees in an ensemble ought to be pruned.  Despite the fact that default constructions of random forests use near full depth trees in most popular software packages, here we provide strong evidence that tree depth should be seen as a natural form of regularization across the entire procedure.  In particular, our work suggests that random forests with shallow trees are advantageous when the signal-to-noise ratio in the data is low.  In building up this argument, we also critique the newly popular notion of ``double descent" in random forests by drawing parallels to U-statistics and arguing that the noticeable jumps in random forest accuracy are the result of simple averaging rather than interpolation.
\end{abstract}


\spacingset{1.25}

\section{Introduction}
\label{sec:intro}

Modern machine learning procedures like neural networks and random forests (RFs) continue to grow in popularity among applied researchers who place a priority on predictive accuracy.  Random forests in particular remain in widespread use as a result of their well-earned reputation for robust, off-the-shelf utility (see e.g.\ \cite{Fernandez2014}) as well as the availability of \emph{ad hoc} tools for assessing variable importance.  As was noted in a recent review paper by \cite{Biau2016}, however, for more than a decade following their introduction, little was known or even formally hypothesized regarding which aspects of the RF procedure were driving their empirical success.

Recently, \cite{Wyner2017} suggested that the reason for the practical success of random forests was interpolation -- a kind of purposeful overfitting.  In the same spirit, \cite{Belkin2019} put forth the more general and now very popular idea of the ``double descent" risk curve in which model error, when plotted against model complexity, exhibits the classical U-shaped curve followed by a second descent beyond the interpolation threshold, indicating that once the model becomes over-parameterized, performance can sometimes be further improved.  The authors provide empirical evidence for this kind of effect with several different models including both neural networks and random forests.  Since then, much effort has been made to provide the theoretical underpinnings for this kind of effect with more tractable models such as ``ridgeless" least squares regression \citep{Hastie2019} and random features regression \citep{mei2019generalization} to name just a few.

An explanation along these lines, however, begs a number of interesting questions.  First, how is model ``complexity" best understood?  Is an average over multiple models really more ``complex" than the individual models themselves?  Second, the idea of relating RF performance to interpolation creates a kind of chicken-and-egg problem:  are RFs accurate \emph{because} they interpolate, or is the (near) interpolation sometimes seen in RFs an innocuous side effect of accurate models constructed in this greedy manner?

More recently, \cite{mentch2020randomization} offered an alternative explanation for RF success from a degrees-of-freedom perspective.  Noticing that interpolation is exceptionally unlikely when RFs are constructed with bootstrap samples and/or used for regression, the authors argue instead that the additional randomness introduced by random feature selection provides a form of implicit regularization. That additional randomness therefore helps to mitigate overfitting, leading to models that substantially outperform bagging alone in low signal-to-noise ratio (SNR) settings, but that lose their advantage at high SNRs. Moreover, the random-feature-subsetting idea was shown to improve many kinds of ensembles with base learners constructed in a greedy fashion -- not just tree-based RFs.  In particular, when the trees in RFs are replaced by linear models, this regularization effect can be proved formally \citep{mentch2020randomization,Lejeune2020implicit}.

 In this work, we adopt this latter regularization framework to investigate the impact of pruning trees in a random forest.  Studies of this sort date back to more than two decades ago when \cite{Breiman1996} first proposed the bagging procedure and showed that aggregating unstable trees built on bootstrap samples can lead to substantial gains in accuracy.  Since then, two separate schools of thought seem to have emerged with one group arguing that trees in an ensemble should be grown full-depth and the other maintaining that tree-depth itself should be seen as a tuning parameter with the potential to greatly impact overall performance.  Indeed, in the original RF proposal by \cite{Breiman2001} as well as in the more recent investigative articles \citep{Wyner2017,Belkin2019,mentch2020randomization}, the trees in RFs are deep (at least near-full depth) and unpruned.  This also remains the standard recommendation in many textbooks (e.g.\ \cite{James2013introduction, Izenman2008modern}) and is the default setting in many standard packages such as \texttt{Scikit-} \texttt{learner} in \texttt{python} and \texttt{randomForest} in \texttt{R}.  On the other hand, there is a wide body of existing work \citep{segal2004machine, Lin2006, Duroux2018, Probst2019} offering strong empirical evidence that proper tuning can significantly improve performance.

Rather than merely adding yet another paper to the growing stack of literature on the topic, this work builds upon the recent insights in \cite{mentch2020randomization} to try and characterize both \emph{why} and \emph{when} pruning trees within a RF has a significant impact on model fit. In particular, we argue that in addition to the bagging (resampling and averaging) and random-feature-subsetting aspects of RFs, tree depth can best be viewed as merely an additional form of regularization, making RFs with shallow trees preferable in low signal-to-noise ratio (SNR) settings.  Ironically, this characterization actually lends support to both schools of thought described above:  at medium to high SNR settings, practitioners are unlikely to notice any meaningful gains in accuracy as a result of pruning individual trees because the bagging and random-feature-subsetting components of the model have already regularized the procedure to a sufficient degree.  On the other hand, when the SNR is very low, pruning trees can offer a third additional form of regularization that can indeed result in improved performance. 

The remainder of the paper is laid out as follows. In Section \ref{sec:Notations}, we formalize the RF notation used throughout the remainder of the paper and offer a more thorough discussion on the previous work in the tree-pruning literature.  A case study on the MNIST dataset designed to replicate the experiments in \cite{Belkin2019} and highlight the shortcomings of the double descent argument is given in Section \ref{sec:RFandDD}.  In Section \ref{sec:RFandTree} we provide both a theoretical motivation and a number of wide-ranging simulations to demonstrate the effect of tree depth on RF performance and properly characterize the settings in which pruning is advantageous.  We conclude with a discussion in Section \ref{sec:discussion}.

\section{Notation and Background}
\label{sec:Notations}

Throughout the remainder of this paper we assume we have a training set $\mathcal{D}_n = \{(\mathbf{X}_1, Y_1), \dots, (\mathbf{X}_n, Y_n)\}$ containing $n$ i.i.d. observations where $Y_i$ denotes the response and $\mathbf{X}_i$ denotes a vector of $p$ features $(X_1, ..., X_p)$. To construct a tree, we begin by resampling $a_n \leq n$ observations from $\mathcal{D}_n$ with or without replacement.  The original random forest formulation utilized bootstrapping so that $a_n=n$ and the sampling is done with replacement, though a number of recent theoretical advances have been made by instead considering subsampling (without replacement) with $a_n=o(n)$ (see e.g., \cite{Scornet2015,Mentch2016,Mentch2017,Wager2018}).  At each step, $\texttt{mtry}\leq p$ eligible features are selected uniformly at random, among which the optimal split is obtained by maximizing the CART criterion \citep{CART}.  The tree continues to split until the number of observations in each cell is less than the pre-specified \texttt{nodesize} or whenever the number of terminal nodes (leaves) reaches \texttt{maxnodes}. In the case of regression, to obtain the prediction at any given point $\bm{x}$, the response values are averaged across all observations that fall into the same leaf as $\bm{x}$.  To form a random forest, the procedure is repeated $B$ times and the final prediction is simply the average across all $B$ tree-level predictions.  In classification settings, the standard approach is to form final estimates via a majority vote at both the tree and forest level.  Algorithm \ref{alg:RF} provides a more detailed summary of this process, which follows closely to Algorithm 1 in \cite{Biau2016}.  Unfamiliar readers are invited to see \cite{Biau2016} for a more detailed discussion.

\begin{algorithm}[!ht]
	\caption{Breiman's (regression) random forest}
	\label{alg:RF}
	\begin{algorithmic}
	\Require Training set $\mathcal{D}_n$, number of trees $B>0$, $a_n \in \{1, \dots, n\}$, \texttt{mtry}$\in \{1,\dots, p\}$, \texttt{nodesize}$\in \{1, \dots, a_n\}$, and $\bm{x} \in \mathcal{X}$
	\Ensure Prediction of the random forest at $\bm{x}$
	\For{$b = 1,\dots,B $}  
			\State Select $a_n$ points, with (or without) replacement, uniformly in $\mathcal{D}_n$. In the following steps, only these $a_n$ observations are used.
			\State Set $\mathcal{P} = (\mathcal{X})$ the list containing the cell associated with the root of the tree.
			\State Set $\mathcal{P}_{\text{final}} = \emptyset$ an empty list.
			\While{ $\mathcal{P} \neq \emptyset$}
				\State Let $A$ be the first element of $\mathcal{P}$.
				\If{A \textit{contains less than} \texttt{nodesize} \textit{points or if all} $\bm{X}_i \in A$ \textit{are equal} }
					\State Remove the cell $A$ from the list $\mathcal{P}$.
					\State $\mathcal{P}_{\text{final}} \leftarrow Concatenate(\mathcal{P}_{\text{final}}, A)$.
				\Else
					\State Select uniformly, without replacement, a subset $\mathcal{M}_{\text{try}} \subset \{1, \dots, p\}$ of cardinality \texttt{mtry}.
					\State Select the best split in $A$ by optimizing the CART-split criterion along the coordinates in $\mathcal{M}_{\text{try}}$.
					\State Cut the cell $A$ according to the best split. Call $A_L$ and $A_R$ the two resulting cells.
					\State Remove the cell $A$ from the list $\mathcal{P}$.
					\State $\mathcal{P} \leftarrow Concatenate(\mathcal{P}, A_L, A_R)$.
				\EndIf
			\EndWhile
			
			\State Compute the predicted value of the $b^{\text{th}}$ tree at $\bm{x}$ equal to the average of the $Y_i$ falling in the cell of $\bm{x}$ in the partition $\mathcal{P}_{\text{final}}$.
			
		\EndFor
		\State Compute the random forest estimate at the query point $\bm{x}$.
	\end{algorithmic}
\end{algorithm}

\subsection{Previous Work on Tree Depth}
When constructing stand-alone decision trees, it has long been understood that the depth of the tree must be carefully chosen to avoid under- or over-fitting.  This is typically accomplished via a cost-complexity parameter that serves to trade off bias and variance by successively pruning away splits in a fully-grown tree whenever the gain in accuracy realized by including them fails to exceed some predefined threshold.  In the case of tree-based ensembles, however, best practices are far less clear and the issue of whether tree depth should be tuned has been the subject of some debate for the past two decades.  
 
Interestingly, Leo Breiman himself -- who proposed both bagging \citep{Breiman1996} and random forests \citep{Breiman2001} -- seemed to flip-flop on this issue.  In the original paper on bagging, \cite{Breiman1996} proposed the idea of best pruned classification and regression trees to be used in the ensemble.  In proposing random forests, however, his advice switched:  ``\emph{Grow the tree using CART methodology to maximum size and do not prune}" \citep{Breiman2001}. 

Many textbooks agree with Breiman's latter advice of constructing RFs with full-depth trees. For example, \cite{Izenman2008modern} states ``\emph{there are only two tuning parameters for a random forest: the number of variables randomly chosen as a subset at each node and the number of bootstrap samples. ...  grow the tree to a maximum depth with no pruning}". Similarly, \cite{James2013introduction} say that ``\emph{To apply bagging to regression trees ... These trees are grown deep, and are not pruned}".  Likewise, the default settings of many widely used statistical computing packages also follow this suggestion of constructing trees to (at least near) full depth. The \texttt{randomForest} package in \texttt{R} constructs trees to the maximum possible depth subject to the constraint of \texttt{nodesize} = 5 for regression and \texttt{nodesize} = 1 for classification trees. Similarly, with \texttt{Scikit-learner} in \texttt{python}, cells are split until all leaves are pure or contain fewer observations than \texttt{min\_sample\_split}, which is set equal to 2 by default.

\begin{figure*}
	\centering
	\includegraphics[width=0.8\textwidth]{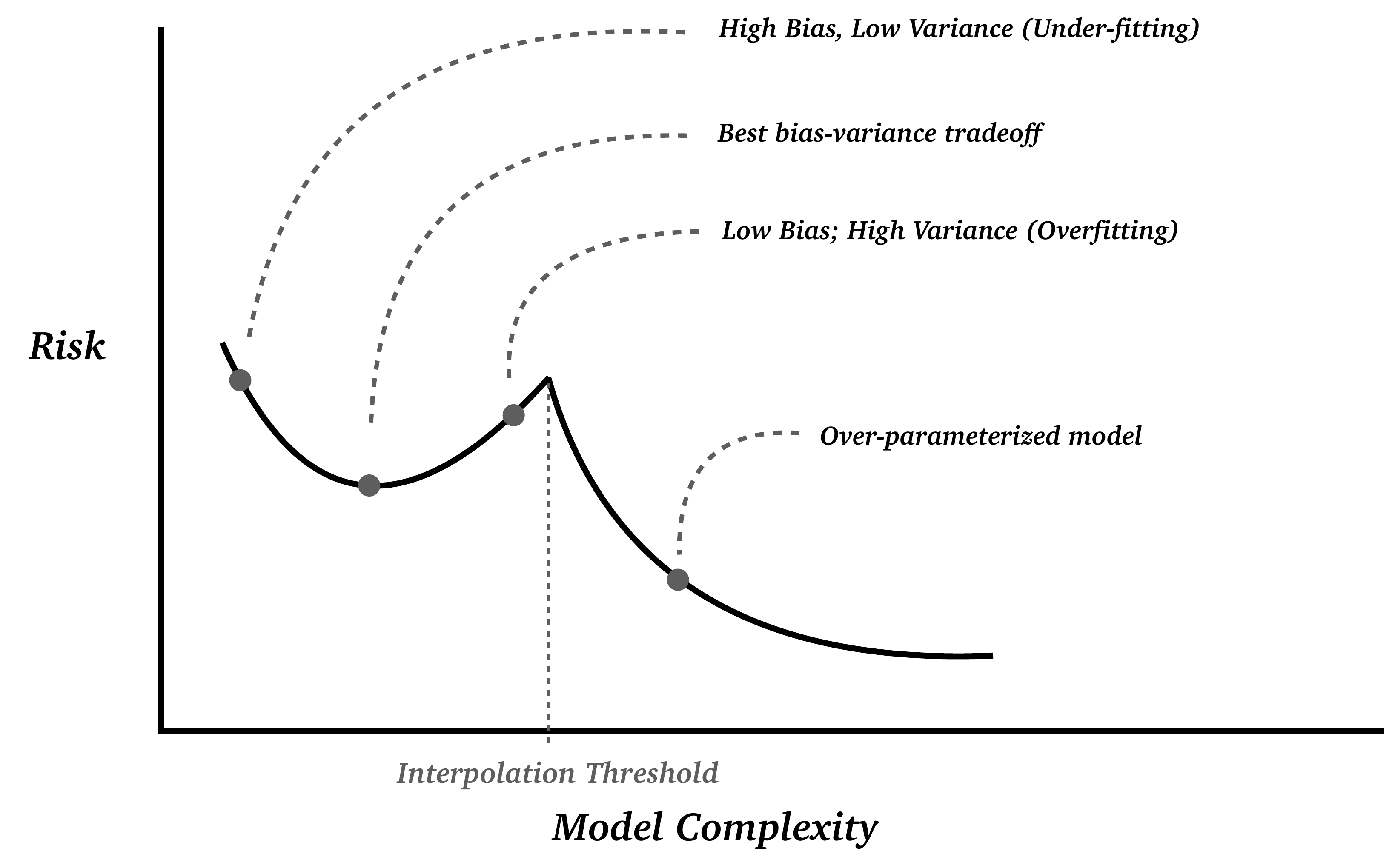}
	\caption{Graphical visualization of the ``double descent" proposed in \cite{Belkin2019}.  Before the interpolation threshold we see the classical U-shaped bias-variance tradeoff, followed by a second descent when considering over-parameterized models. }
	\label{fig:InterpPlot}
\end{figure*}

Despite the fact that building trees in RFs to full-depth has largely become the standard advice offered, the fact remains that numerous studies have provided strong empirical evidence that tuning tree depth can substantially impact performance.  \cite{segal2004machine} performed a number of experiments on both synthetic and real-world data utilizing various tree depths and feature subsampling rates and found significant changes in accuracy, but no clear trends in terms of which settings were optimal in which settings.  \cite{Lin2006} put forth the idea of adaptive nearest neighbors to better characterize the class of models to which RFs belong and, as related to the idea of tree depth, argued that ``\emph{growing the largest tree was not optimal in general}". \cite{Duroux2018} carried out simulations directly comparing the performance of Breiman's original RFs to RFs constructed with shallow trees and reached much the same conclusion.  The popular graduate-level textbook ``The Elements of Statistical Learning" \citep{esl} takes more of a neutral, measured approach, saying only that ``\emph{the average of fully grown trees can result in too rich a model}" but that ``\emph{Our experience is that using full-grown trees seldom costs much, and results in one less tuning parameter}".  

Despite the substantial amount of previous literature on the topic, to the best of our knowledge, none of this work has offered a principled explanation as to \emph{why} limiting tree-depth in RFs could be helpful or \emph{when} such an alternative construction might be expected to outperform Breiman's original proposal.  In the following sections, we aim to take a step forward in this direction by building upon the regularization framework suggested in \cite{mentch2020randomization}.  To begin, we first revisit the double descent argument put forth in \cite{Belkin2019} and demonstrate that in the context of random forests, this second drop in risk is not only achievable, but expected, even when shallow decision trees are employed and interpolation is exceedingly unlikely.

\section{Random Forests and Double Descent}
\label{sec:RFandDD}

In work that has since gained a lot of attention, \cite{Belkin2019} studied the the relationship between performance (risk) and complexity for a variety of supervised learning models.  Classical statistical theory suggests that such a curve should exhibit a natural U shape:  models with complexity too low will have too much bias and under-fit, models with too much complexity will have too much variance and over-fit, and thus the optimal model that minimizes this curve should be that which optimally trades off bias and variance.  Much to the surprise of many in the statistics community, however, \cite{Belkin2019} noticed that for a variety of such models, the risk curve can begin a second descent once the complexity crosses beyond the interpolation threshold and becomes over-parameterized -- see Figure \ref{fig:InterpPlot}.  In some cases, this second descent achieves minimum values that exceed even those obtained by the model believed to be optimal by classical theory -- that which minimizes the first descent.  Put simply, the work suggested that the poor performance often seen in parameter-rich, highly-complex models could be mitigated not only by removing parameters, but also by adding them.  \cite{Belkin2019} attributed this second descent to the argument that despite the large model complexity of interpolating functions, their function space norm is smaller and thus they can be seen as ``simpler" though an alternative lens.

Before continuing, it is worth pausing to more carefully define the notions of interpolation and model complexity as these are clearly of fundamental importance to the argument put forth in \cite{Belkin2019}.  Interpolating functions are simply those whose training error is zero, as stated more explicitly in Definition \ref{def}. 

\begin{definition}
\label{def}
	A learning model $\hat{f}$ is an interpolating function if for every training point $(\bm{X}_i, Y_i) \in \mathcal{D}_n$, $\hat{f}(\bm{X}_i) = Y_i$. 
\end{definition}

\noindent The notion of interpolation is straightforward, easily defined, and can pertain to both classification and regression problems.  The interpolation threshold is then identified as the minimum complexity at which the model begins to interpolate. 

The idea of model \emph{complexity}, on the other hand, is far more nuanced.  \cite{Belkin2019} define the complexity of a model as ``\emph{the number of parameters needed to specify a function within the class}". Such a definition is natural and intuitive for simple models like ordinary least squares (OLS) linear regression or individual decision trees and is consistent with notion of \textit{the effective number of parameters} discussed in \cite{esl}.  However, this notion of complexity becomes less straightforward for black-box models.  With random forests, for example, \cite{Belkin2019} applies this definition in something of a hybrid fashion:  for random forests containing only a single tree, complexity is taken as the number of leaves so that the original (training) sample size $n$ serves as an upper bound.  Beyond this point, however, the complexity of ensembles of full-depth trees seems to be measured as (at least proportional to) the number of trees in the ensemble.  This is where the natural intuition behind such a definition begins to break down.  While averaging $B$ trees can potentially partition the feature space in a finer way (and thus utilize more ``parameters") relative to a forest with $B-1$ trees, it seems a bit unorthodox to assume that the former estimator is inherently more ``complex". 

A natural analogy can be drawn here to classical U-statistics \citep{HoeffdingUstat}.  Recall that the standard motivation for such estimators is as follows.  Given a sample $Z_1, ..., Z_n$ of size $n$ and a parameter of interest $\theta$, we assume there exists some unbiased estimator $h(Z_1, ..., Z_k)$ utilizing only $k\leq n$ arguments and that $h$ is permutation symmetric in those $k$ arguments.  This base estimator $h$ is generally referred to as a kernel of rank $k$.  While any subsample of size $k$ from the original sample will suffice to produce an unbiased estimate of $\theta$, not surprisingly, a better estimator, and indeed, that which has the minimum variance, is the U-statistic
\[U_n = \frac{1}{\binom{n}{k} } \sum_{(n,k)} h(Z_{1}^{*}, ..., Z_{k}^{*})\]
formed by evaluating $h$ over all $\binom{n}{k}$ possible subsamples and averaging.  When $B < \binom{n}{k}$ subsamples are utilized, the resulting estimator is referred to as an incomplete U-statistic.

Note that when the individual kernels $h$ are seen as (possibly randomized) decision trees, there is an immediate connection between U-statistics and random forests.  In fact, it was this connection that was exploited by \cite{Mentch2016} to demonstrate that RF predictions are asymptotically normal when trees are constructed via subsampling instead of traditional bootstrap samples.

This connection also helps make clear the shortcomings of the interpolation-based argument for why a second descent is observed when random forests begin consisting of more than one tree.  Decades-old statistical theory tells us that U-statistic-style estimators have the same bias as the original estimator $h$ but with smaller variance and would thus be preferred according to the classic bias-variance tradeoff.  By near-identical reasoning, RFs containing many trees built with resamples of the original data ought to be preferred to individual decision trees.  Seen in this fashion, it does not necessarily make intuitive sense to define one such estimator as more ``complex" than another merely because it takes an average over a larger collection.

Even more importantly for our purposes here though, estimators constructed by taking a larger average should be preferred \emph{regardless} of the subsample size (rank) $k$.  Just as a U-statistic formulation would be expected to result in an improved estimator regardless of the rank of the kernel, random forests with many trees should be preferable to individual decision trees regardless of the subsample size on which they're built.  That is, even when relatively shallow trees are constructed so that complexity falls well short of the interpolation threshold, a second drop in random forest risk should be expected as trees are added to the ensemble.  In some classification settings or in settings where trees are each constructed on the same original sample it may be possible to interpolate (at least nearly), but interpolation is not the \emph{cause} of this.  In the following subsection, we elaborate on this point by replicating the analysis in \cite{Belkin2019} before arguing that this kind of shallow-tree construction is actually preferable in noisy data settings in Section \ref{sec:RFandTree}.

\begin{figure*}[!ht]
	\centering
	\includegraphics[width = 0.32\columnwidth]{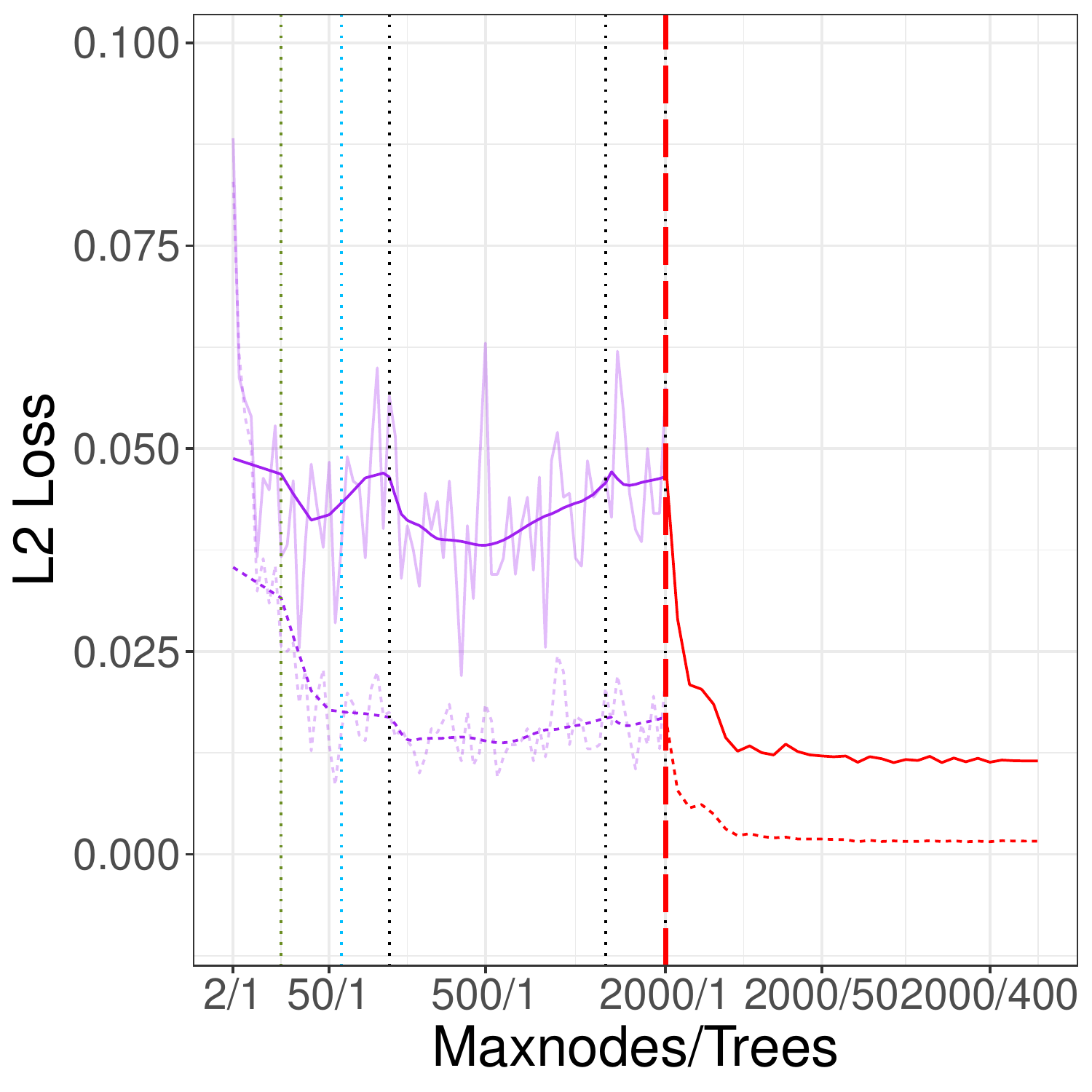}
	\includegraphics[width = 0.32\columnwidth]{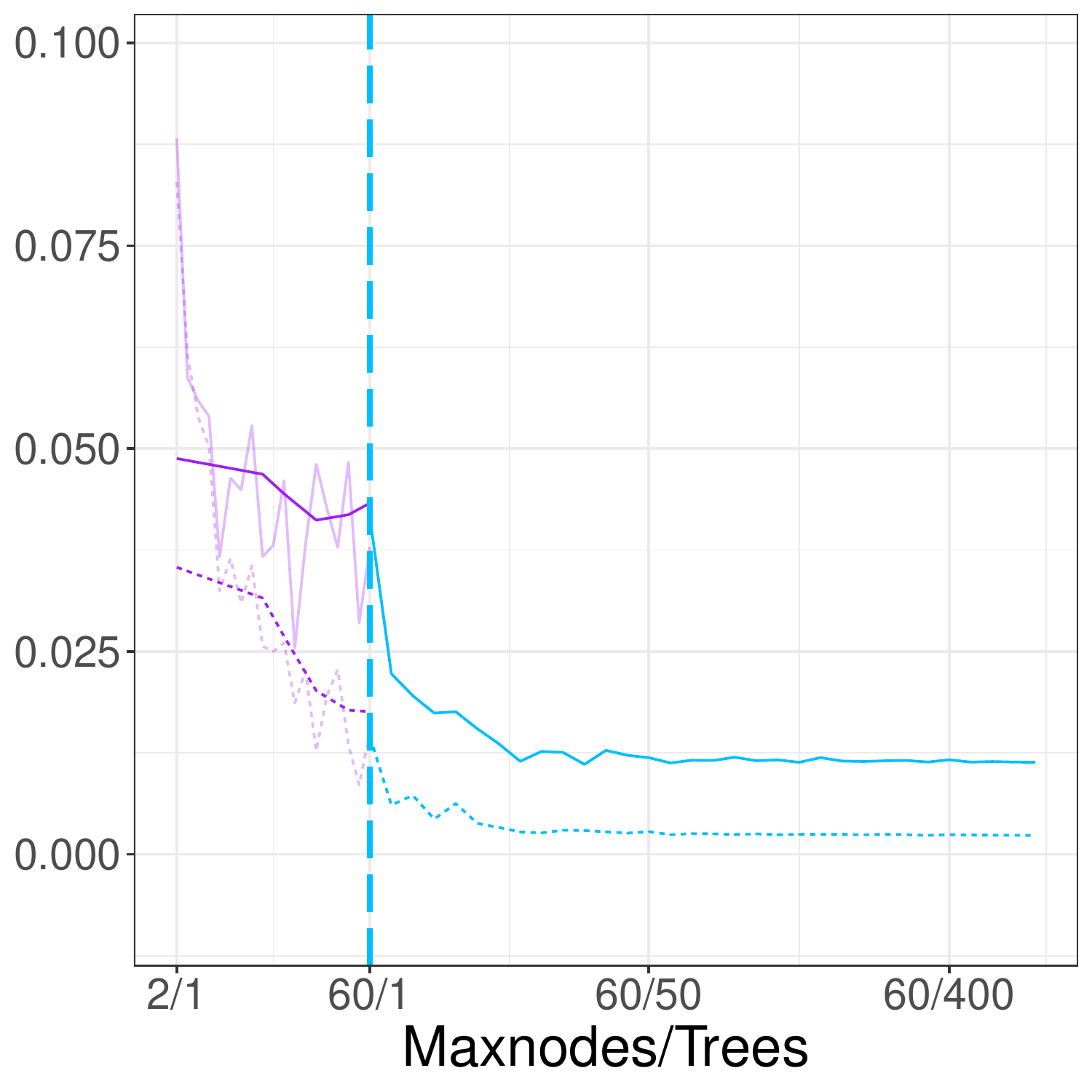}
	\includegraphics[width = 0.32\columnwidth]{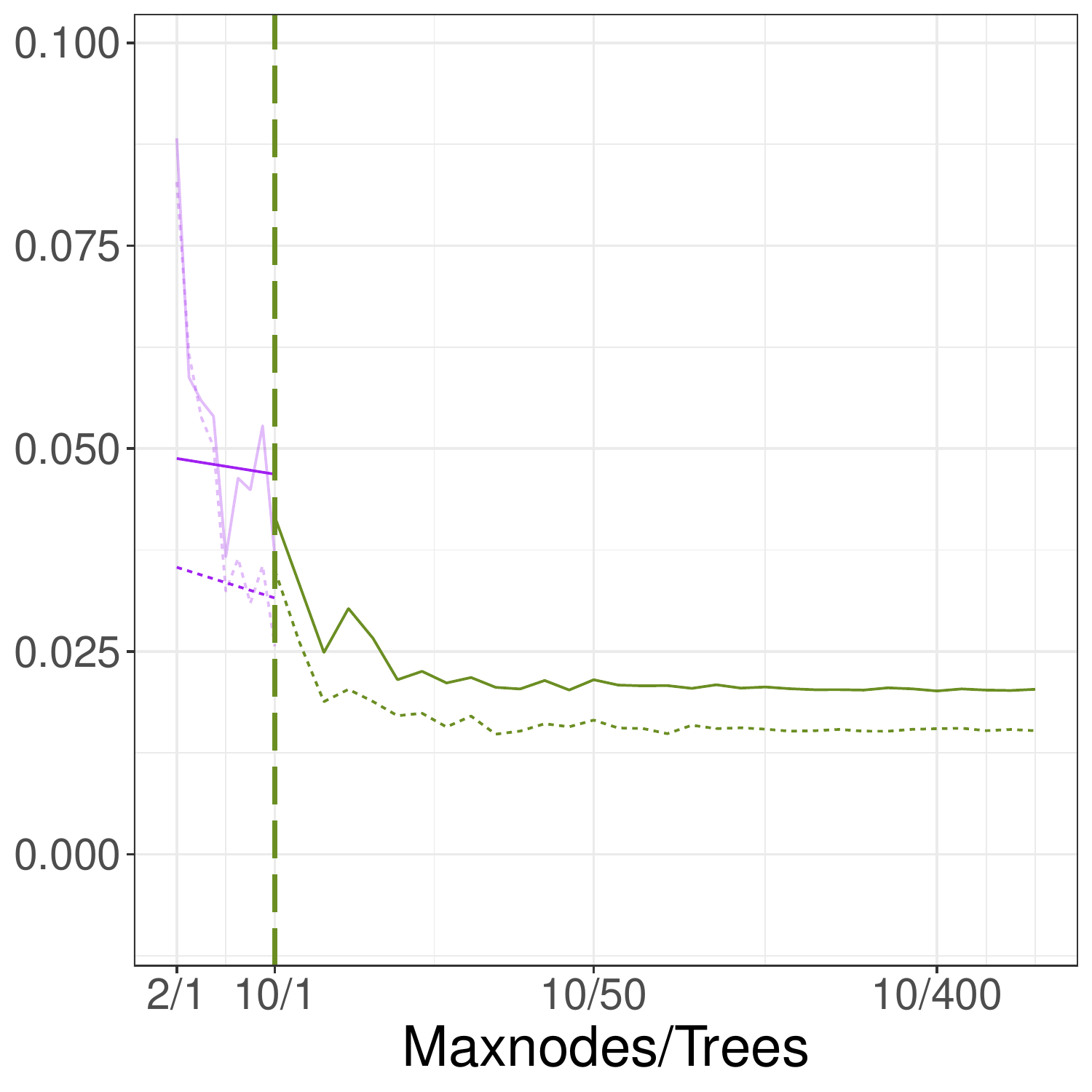}
	\includegraphics[width = 0.32\columnwidth]{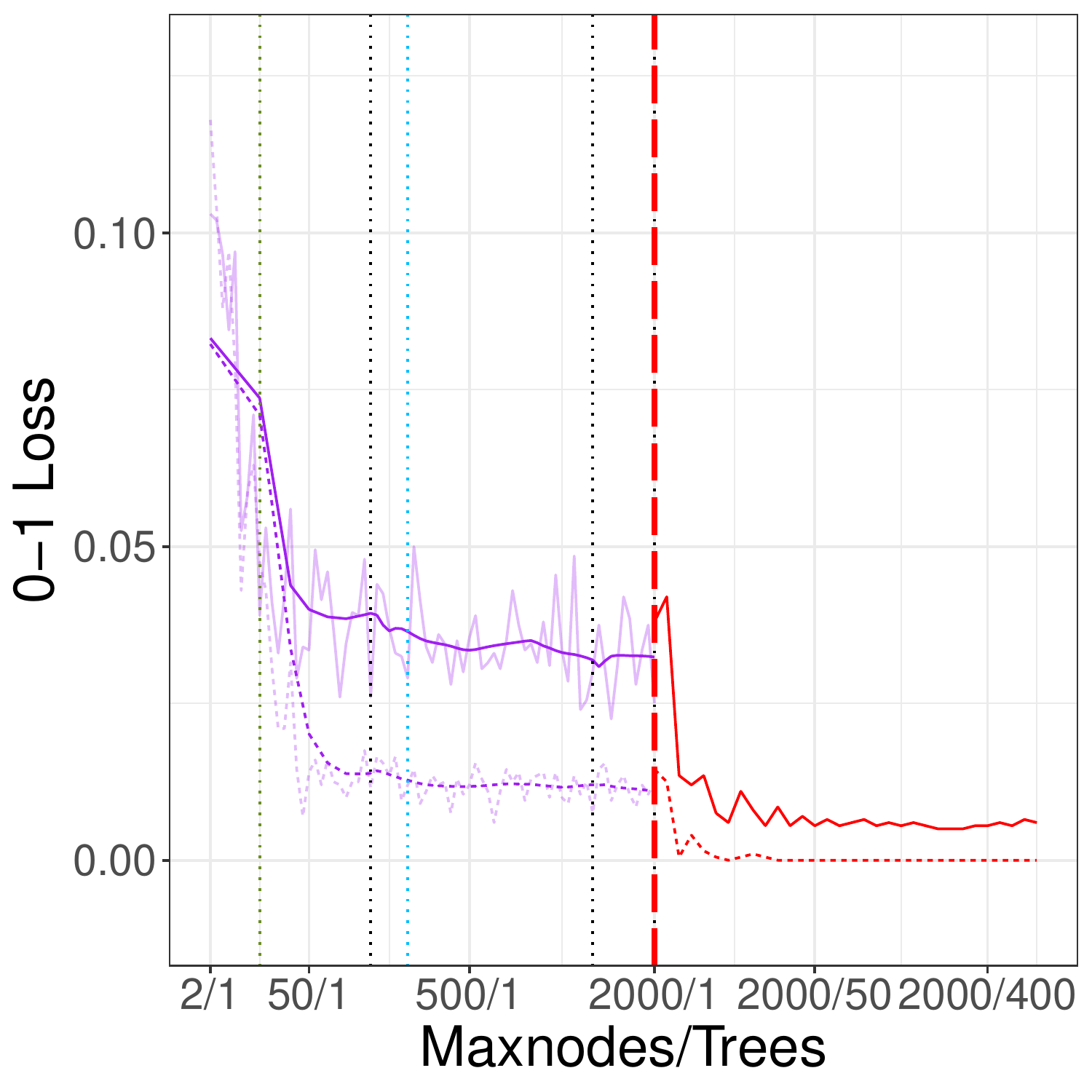}
	\includegraphics[width = 0.32\columnwidth]{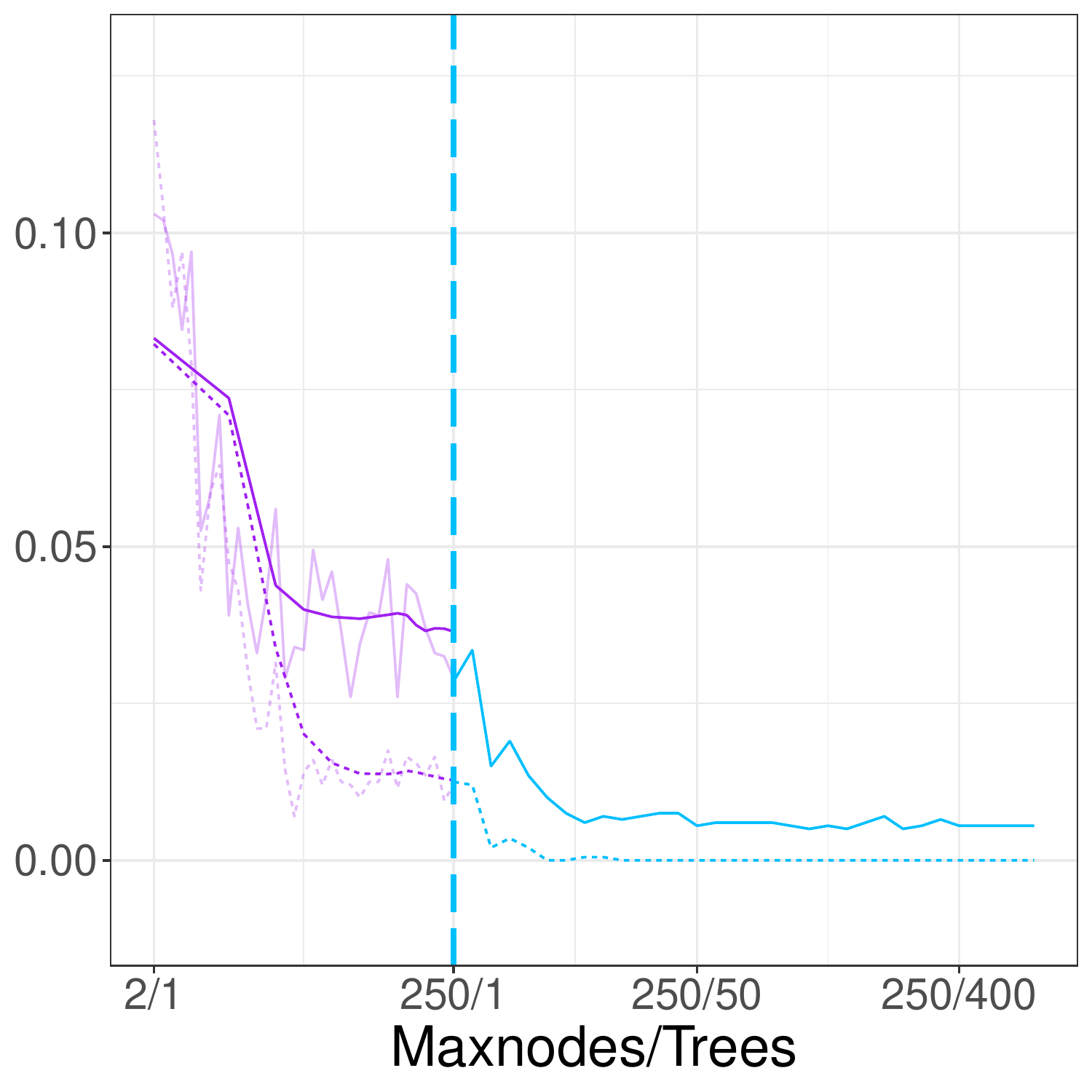}
	\includegraphics[width = 0.32\columnwidth]{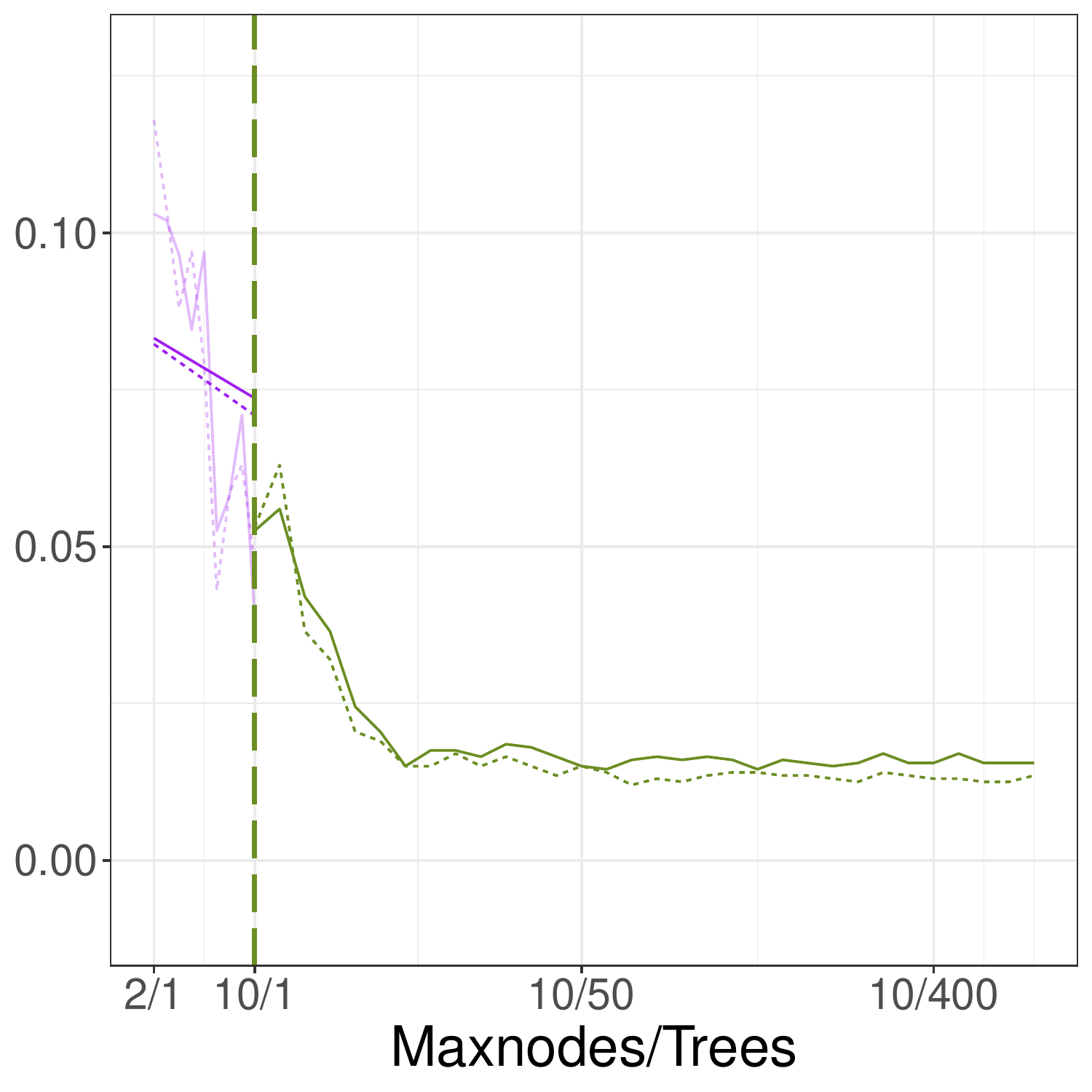}
	\caption{Performance of a single randomized tree (purple curves), full-depth RFs (left column, red curves), tuned RFs (middle column, blue curves) and shallow RFs (right column, green curves) using \textbf{bootstrap samples} in regression (top row) and classification (bottom row) settings. The transparent purple background curve is the raw result of a single randomized tree; the bold purple line corresponds to a lowess smooth. Dashed and solid curves correspond to the performance on training and test sets, respectively.}
	\label{fig:MNIST_boot}
\end{figure*}

\begin{figure*}[!ht]
	\centering
	\includegraphics[width = 0.32\columnwidth]{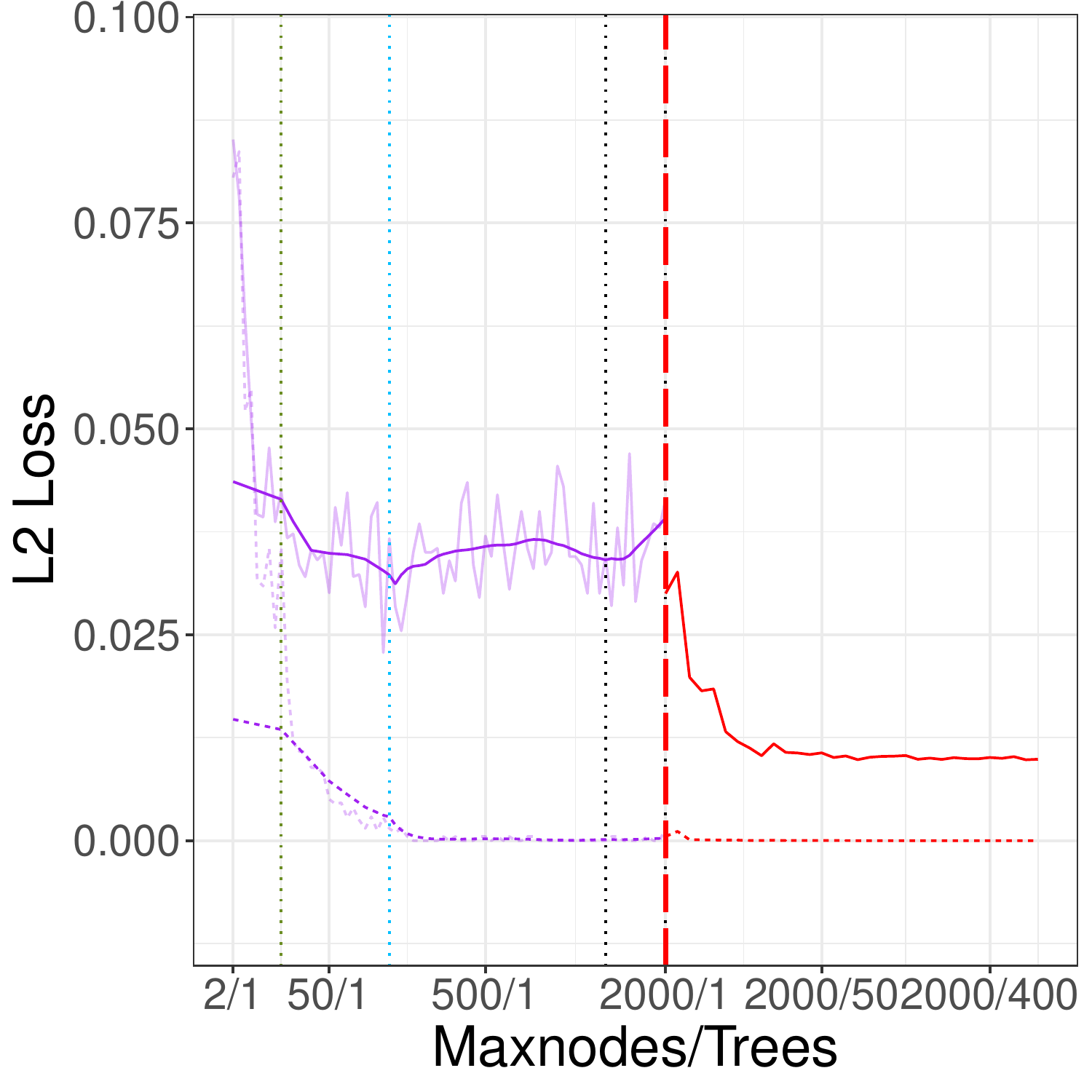}
	\includegraphics[width = 0.32\columnwidth]{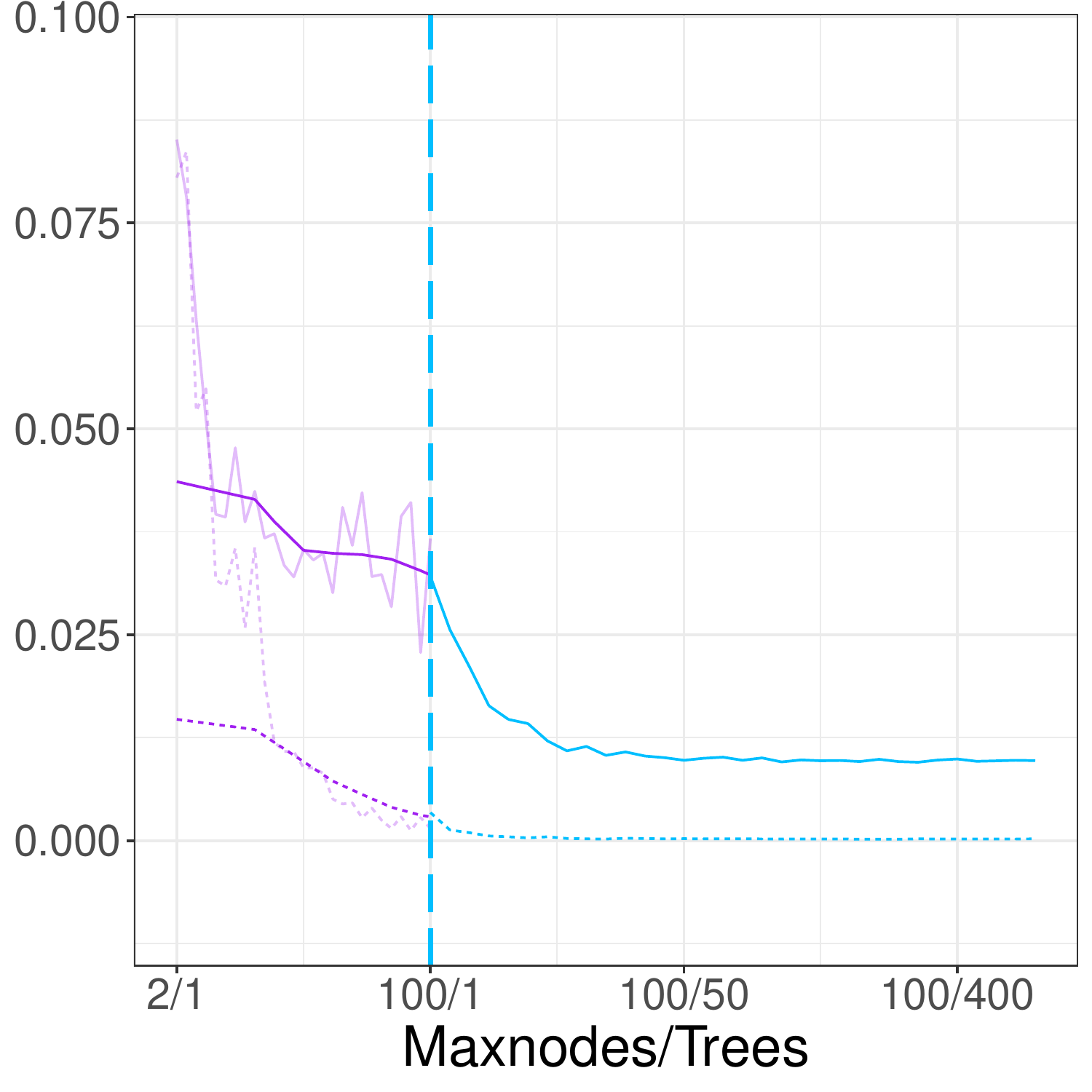}
	\includegraphics[width = 0.32\columnwidth]{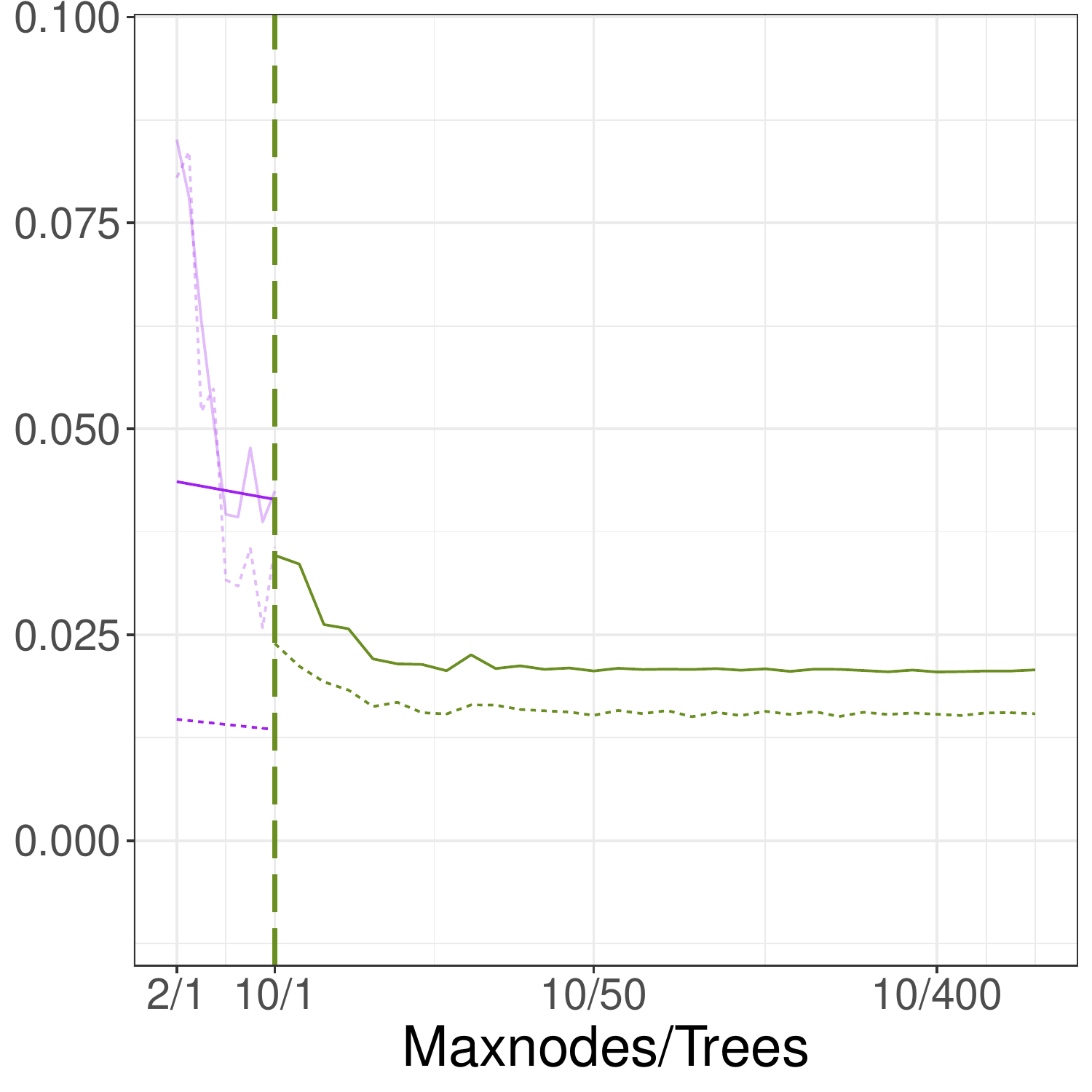}
	\includegraphics[width = 0.32\columnwidth]{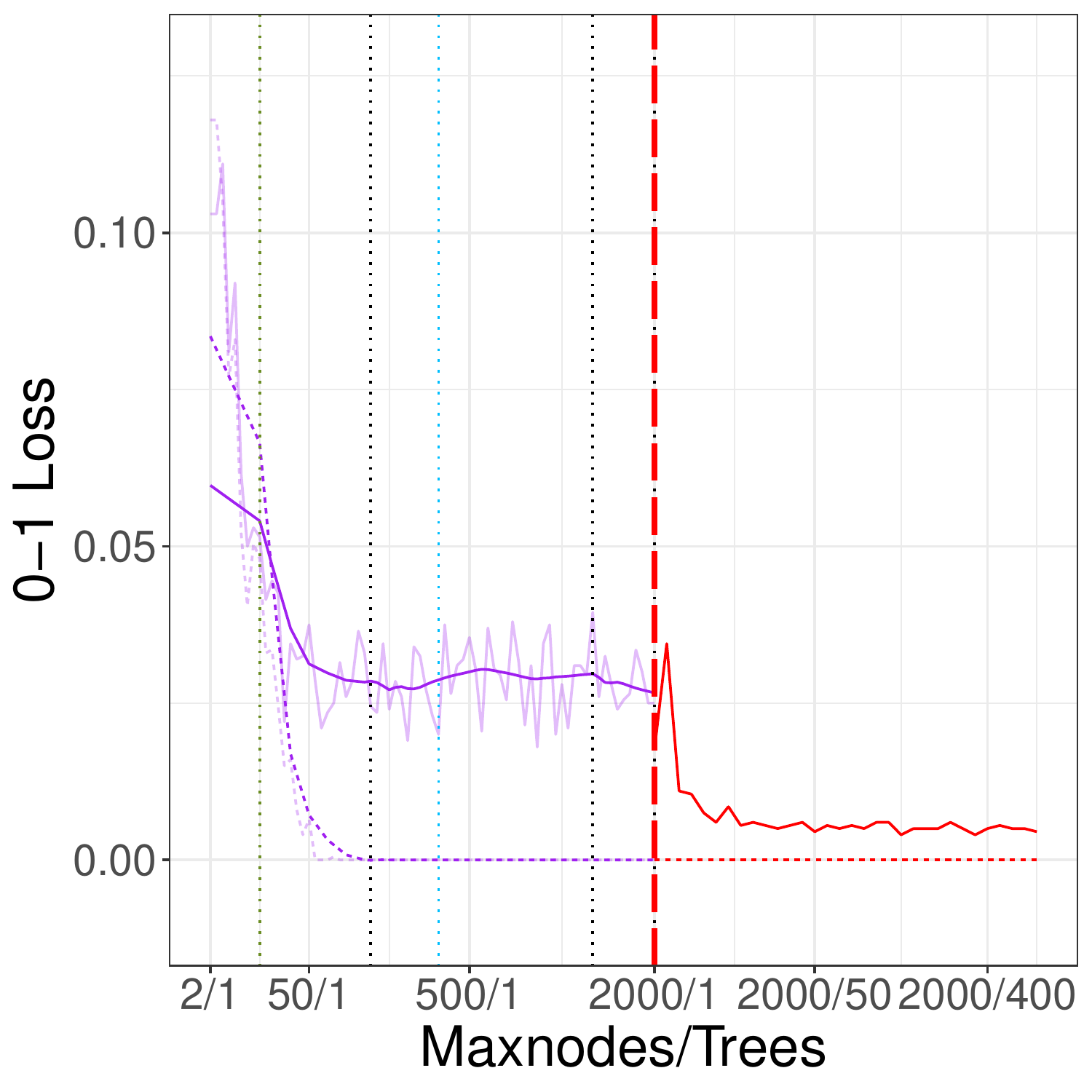}
	\includegraphics[width = 0.32\columnwidth]{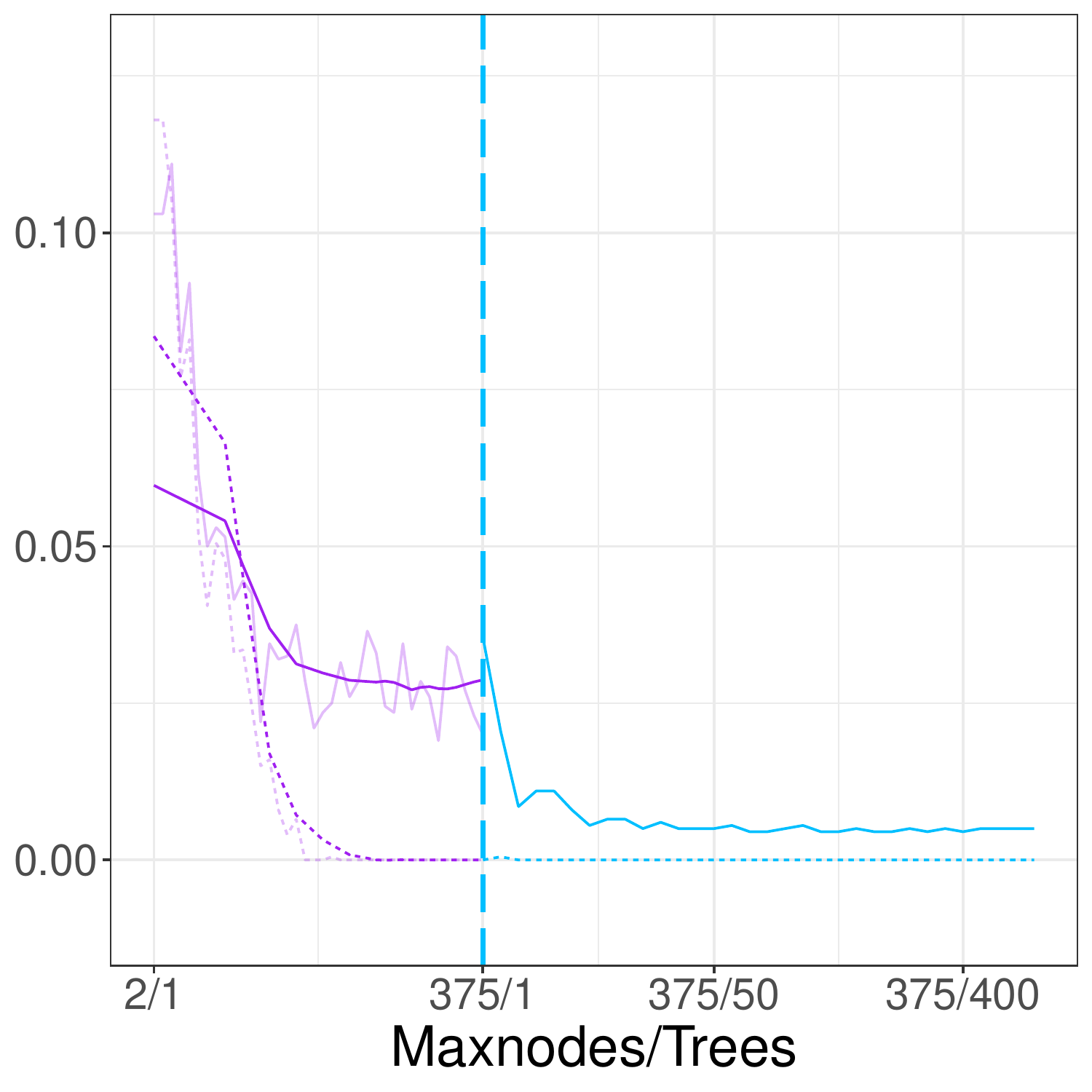}
	\includegraphics[width = 0.32\columnwidth]{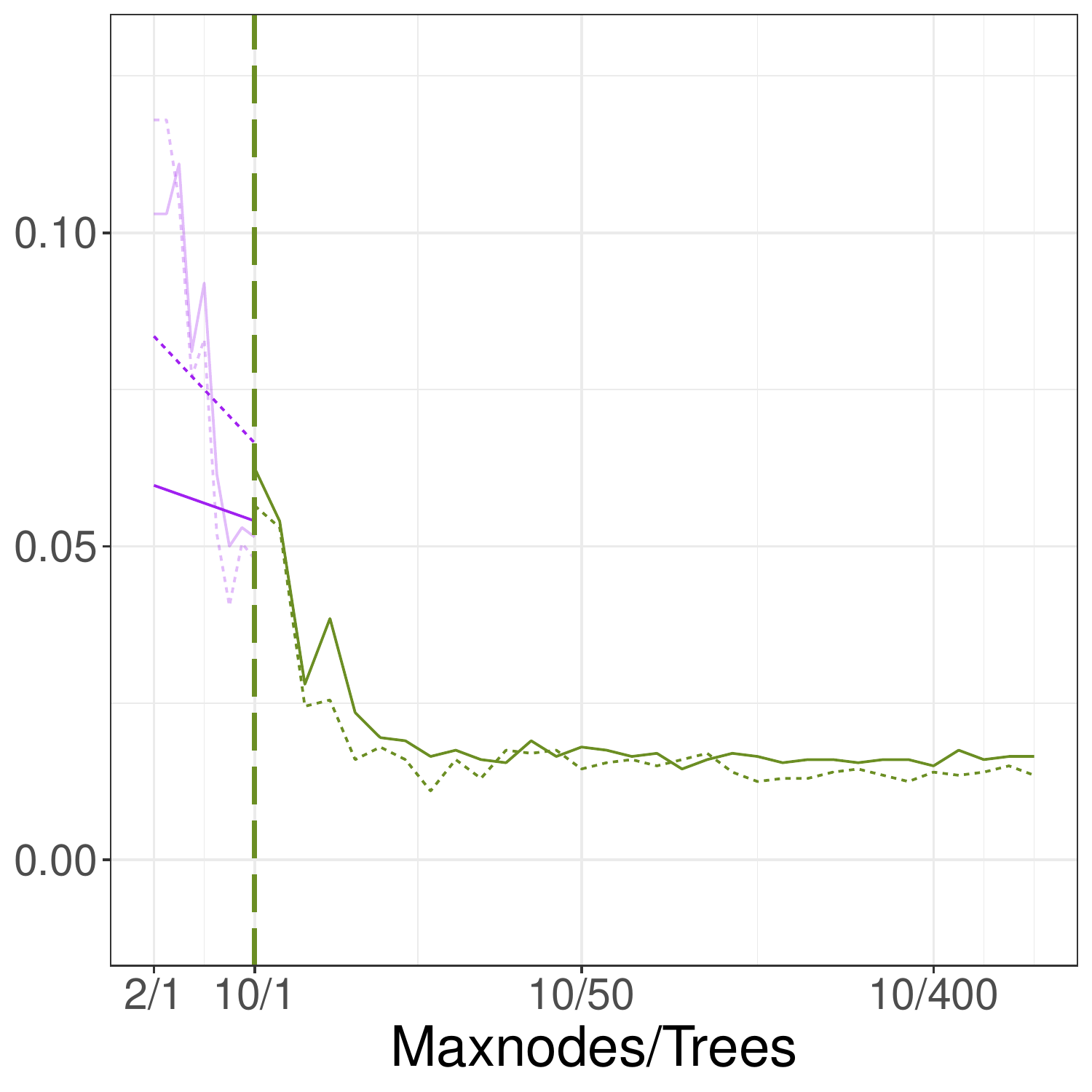}
	\caption{Performance of a single randomized tree (purple curves), full-depth RFs (left column, red curves), tuned RFs (middle column, blue curves) and shallow RFs (right column, green curves) using \textbf{original samples} in regression (top row) and classification (bottom row) settings. The transparent purple background curve is the raw result of a single randomized tree; the bold purple line corresponds to a lowess smooth. Dashed and solid curves correspond to the performance on training and test sets, respectively.}
	\label{fig:MNIST_noboot}	
\end{figure*}

\subsection{A Case Study on the MNIST Dataset}
\label{sec:3.1}

Some of the data employed by \cite{Belkin2019} to demonstrate the double descent curve was from the MNIST database, which contains data (60,000 training samples and 10,000 test samples) extracted from handwritten numeric digits.  In order to replicate these kinds of experiments, we begin by randomly sampling a training set $\mathcal{D}_{\text{Train}}$ from the original training set, as well as test and validation sets $\mathcal{D}_{\text{Test}}$ and $\mathcal{D}_{\text{Val}}$ from the original test set, each of size $n=2000$.  As the response, we use a binary indicator corresponding to whether or not the handwritten digit is a `1'. 

In this subsection, we refer to the model complexity of RFs as tree depth/number of trees so as to be consistent with definitions and plots in \cite{Belkin2019}. Tree depth is controlled by two parameters, \texttt{maxnodes} and \texttt{nodesize}.  The \texttt{maxnodes} parameter can range from 2 (a stump) to 2000 (one observation in each leaf) and we set the \texttt{nodesize} parameter to 1 so that an internal node will be continue to be split until the number of leaves is equal to \texttt{maxnodes}.

We consider a total of four different model setups: 
\begin{itemize}[label=\textbullet]
	\item \textbf{Individual Trees: } A single randomized tree with the number of \texttt{maxnodes} ranging from 2 to 2000
	\vspace{1mm}
	\item \textbf{Full-depth Forests: } 500 trees constructed with the \texttt{maxnodes} set equal to 2000 so that all trees are full depth.
	\vspace{1mm}
	\item \textbf{Tuned Forests: } 500 trees constructed with \texttt{maxnodes} set equal to the depth \texttt{maxnodes}$_{opt}$ that minimizes the error of a single (randomized) tree on the validation set.
	\vspace{1mm}
	\item \textbf{Shallow Forests: } 500 trees constructed with \texttt{maxnodes} = 10.
\end{itemize}

\noindent Note that each setup utilizes randomized trees as in a typical random forest.  For each setup, we consider constructing trees via bootstrap samples (as traditionally done) and also by using the same original training set each time (as was done in \cite{Belkin2019}). All remaining parameters are set to the default values in the \texttt{R} package \texttt{randomForest}.  We consider both classification and regression setups.  In the classification setup, both tree- and forest-level predictions are made via majority vote and performance on the test set is measured by 0-1 loss.  In the regression setting, we utilize regression trees so that the binary responses are averaged at both the tree and forest level and performance is measured via the standard $L_2$ (squared error) loss. 

Results from forests constructed with bootstrap samples and original samples are shown in Figures \ref{fig:MNIST_boot} and \ref{fig:MNIST_noboot} respectively. In each figure, the top row corresponds to the regression setup and the bottom to the classification setup.  As in \cite{Belkin2019}, model complexity is shown on the horizontal axis and model performance on the vertical axis with the training loss and test loss represented by dashed and solid lines, respectively.  The four colors in the plots represent the performance of the four setups under consideration:  purple for individual trees, red for full-depth RFs (left column), blue for tuned RFs (middle column), and green for shallow RFs (right column). Randomization in the single tree leads to a great deal of variation in the performance so at the level of individual trees, we also include a bolder purple line corresponding to a lowess smooth. The bold vertical long-dashed lines in each plot denote the transition points from a single tree to a RF with multiple trees.  In the left-most column of each figure, there are additional black vertical dotted lines in each plot that are included to indicate that model complexity between these lines is equally spaced. 

Plots in the left-most column of each figure are designed to replicate the kind of analysis done in \cite{Belkin2019}.  Here we see that once the interpolation threshold is reached and more trees are added to the forest (red dashed vertical line), the performance of full-depth RFs improves and eventually levels off indicating what \cite{Belkin2019} describe as the second descent.  A close inspection of the upper-left-hand plots in Figures \ref{fig:MNIST_boot} and \ref{fig:MNIST_noboot} is also very revealing.  Note that in Figure \ref{fig:MNIST_boot}, where bootstrapping is employed, the training error never reaches 0 (the RFs do not interpolate) and yet, once more trees are added, a near identical drop in test error is observed compared to the no-bootstrap case (Figure \ref{fig:MNIST_noboot}) where trees are constructed on the same original data each time and 0 training error is achieved quite quickly.  Furthermore, note that tuned RFs (middle column) perform almost identically to their full depth counterparts.  That is, even though the training loss is not 0 at the transition point from single trees to ensembles of trees, a rapid performance improvement is still obvious once more trees added.  Once again, in the right-most plots, the shallow RFs again exhibit a dramatic drop in loss when more trees are added despite the fact that the training error remains well above 0.

These demonstrations make clear that while we do indeed observe a dramatic drop in loss with full-depth trees once multiple trees are considered -- what \cite{Belkin2019} refers to as the ``interpolation threshold" -- near identical performance improvement jumps are seen regardless of the training error whenever averaging over many trees as opposed to considering only a single tree.  Indeed, the idea that averaging (or otherwise aggregating) unstable estimators to reduce the variance of the procedure is not a new idea (see e.g.\ \cite{Breiman1996, buhlmann2002analyzing}) nor is it an idea that is in any way dependent on interpolation of the individual base models.  Nonetheless, when comparing the test performance of shallow RFs (right-most column) to their counterparts in Figures \ref{fig:MNIST_boot} and \ref{fig:MNIST_noboot}, it's clear that these ensembles of shallow learners are not performing as well as the ``deeper" ensembles in the left and middle columns.  One may thus be tempted to argue, as continues to be claimed in many textbooks, that ensembles of deeper trees are still preferable nonetheless.  In the following section, we dispel this idea and demonstrate that in noisy data settings, ensembles of shallow trees are indeed preferable.

\section{Random Forests and Tree Depth}
\label{sec:RFandTree}

We begin our investigation into the relationship between tree depth and RF accuracy by first establishing a theoretical motivation for the idea that tree depth acts as a natural means of regularization. We then conclude with a number of simulations to empirically demonstrate this effect. 

\subsection{Tree Depth as Regularization}
\label{sec:theory}

Recently, \cite{mentch2020randomization} argued that the random-feature-subsetting aspect of random forests implicitly regularizes the RF procedure and produces a kind of shrinkage; when data are more noisy, more randomness (i.e.\ a smaller value of \texttt{mtry}) helps prevent overfitting and should be preferred.  Furthermore, it was shown that this is not a tree-specific property, but instead applies equally well to any ensemble consisting of base models constructed in a greedy fashion. In particular, \cite{mentch2020randomization} proposed a randomized forward selection procedure (RandFS) designed to mimic a traditional RF but with linear model base learners constructed via a kind of randomized forward selection in which only $\texttt{mtry} \leq p$ features (selected uniformly at random) are available to be added at each step.  We now utilize this same setup to demonstrate the regularizing effect of model size (depth) for such randomized ensembles.

Suppose that we construct $B$ linear models in the randomized fashion described above, each to a depth of $d$.  For $b = 1, \dots, B$, let $\hat{f}_{\RFS, d}^{(b)}$ denote the $b^{\text{th}}$ base learner, which is of the form
\begin{align}
	\hat{f}_{\RFS, d}^{(b)} = \hat{\beta}_{0}^{(b)} + \hat{\beta}_{(1)}^{(b)}X_{(1)}^{(b)} + \dots + \hat{\beta}_{(d)}^{(b)}X_{(d)}^{(b)} \label{eqn:RFS}
\end{align}
where $X_{(j)}^{(b)}$ denotes the feature selected at the $j^{\text{th}}$ step in the $b^{th}$ model and $\hat{\beta}_{(j)}^{(b)}$ is the corresponding coefficient estimate.  Similarly, define $\hat{\beta}_{j}^{(b)}$ to be the coefficient estimate for $X_j$ in the $b^{th}$ model which, for example, will be 0 whenever $X_j$ is not selected to be included in that model. To form the ensemble, we average across these $B$ individual linear models to form
\[
	\hat{f}_{\RFS,d} = \frac{1}{B} \sum_{b = 1}^{B} \hat{f}_{\RFS, d}^{(b)} := \hat{\beta}_{0} + \sum_{j = 1}^{p} \hat{\beta}_{j}X_j
\]
where each $\hat{\beta}_k := 1/B \, \sum_{b=1}^{B} \hat{\beta}_{k}^{(b)}$ is simply the average across each individual estimate for $k=0, ..., p$.  Note that even though each individual linear model is of size $d$, the ensemble model will generally contain more than $d$ features since some features may be included in some models but not others.

Now consider a situation where the OLS estimator exists and denote the OLS estimate for the coefficient of $X_j$ by $\hat{\beta}_{j, OLS}$. Given an orthogonal design matrix, the linear model in (\ref{eqn:RFS}) can be rewritten as
\begin{align*}
	\hat{f}_{\RFS, d}^{(b)} = \hat{\beta}_{0, \OLS} + \hat{\beta}_{1, \OLS}I_{1, d}^{(b)} X_1 + \dots +  \hat{\beta}_{p, \OLS}I_{p, d}^{(b)} X_p
\end{align*}
where $I_{j,d}^{(b)}$ is an indicator function set equal to 1 if $X_j$ appears in that $b^{th}$ model and is 0 if $X_j$ is not selected.  Then, we can define $\gamma_{j,d} = 1/B \, \sum_{b = 1}^{B} I_{j, d}^{(b)}$ as the proportion of models in which $X_j$ is selected for each $j=1, ..., p$ and we can write
\begin{align*}
	\hat{\beta}_{j} = \gamma_{j,d} \hat{\beta}_{j, \OLS}  \label{eqn:RFS_beta}.
\end{align*}

Note that for each $j=1, ..., p$, $0 \leq \gamma_{j,d} \leq 1$ and thus, in this context, this process of averaging across randomized linear models is equivalent to constructing a linear model that includes all $p$ of the original covariates, but where each OLS coefficient estimate $\hat{\beta}_{j, \OLS}$ is shrunk by some amount $\gamma_{j,d}$.  Even more importantly for the conversation on model depth, note that the magnitude by which we shrink the coefficient estimates is directly related to the model size (depth) $d$:  as $d$ becomes smaller, each model must include fewer covariates and thus some covariates must necessarily appear in less models, leading to more shrinkage on their corresponding coefficient estimates.  Note also that this shrinkage is not applied uniformly across all covariates; covariates more correlated with the response will have a higher chance of being selected once made available and thus, on average, will be shrunk less than covariates with a weaker relationship to the response.  In these ways, this kind of randomized linear model ensemble behaves similarly to procedures like ridge regression and lasso that take an explicitly penalized approach to fitting.  Recent work in both \cite{Lejeune2020implicit} and \cite{mentch2020randomization} further detail the connections between this kind of RandFS procedure and classical ridge regression, providing conditions under which the two approaches are equivalent.  We refer readers interested in a more in-depth theoretical discussion to those recent studies.

Finally, note that in noisy data settings where the signal is relatively low, this kind of shrinkage can be advantageous by preventing overfitting to noise.  Since RFs are constructed in the same kind of fashion by using trees instead of linear models, we ought to expect the same kind of outcome:  ensembles of shallow trees ought to perform better in noisier settings. We now explore this empirically via a number of simulations in the following subsection.

\begin{figure*}[!ht]
	\centering
	\includegraphics[width = 0.3\columnwidth]{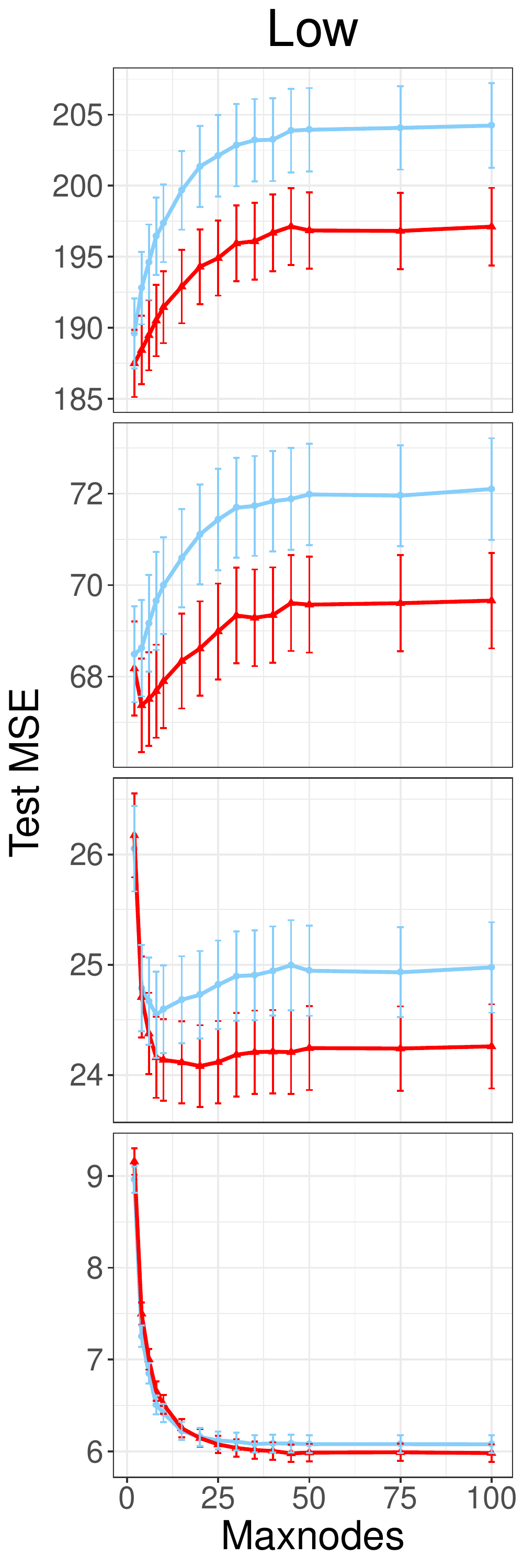}
	\includegraphics[width = 0.3\columnwidth]{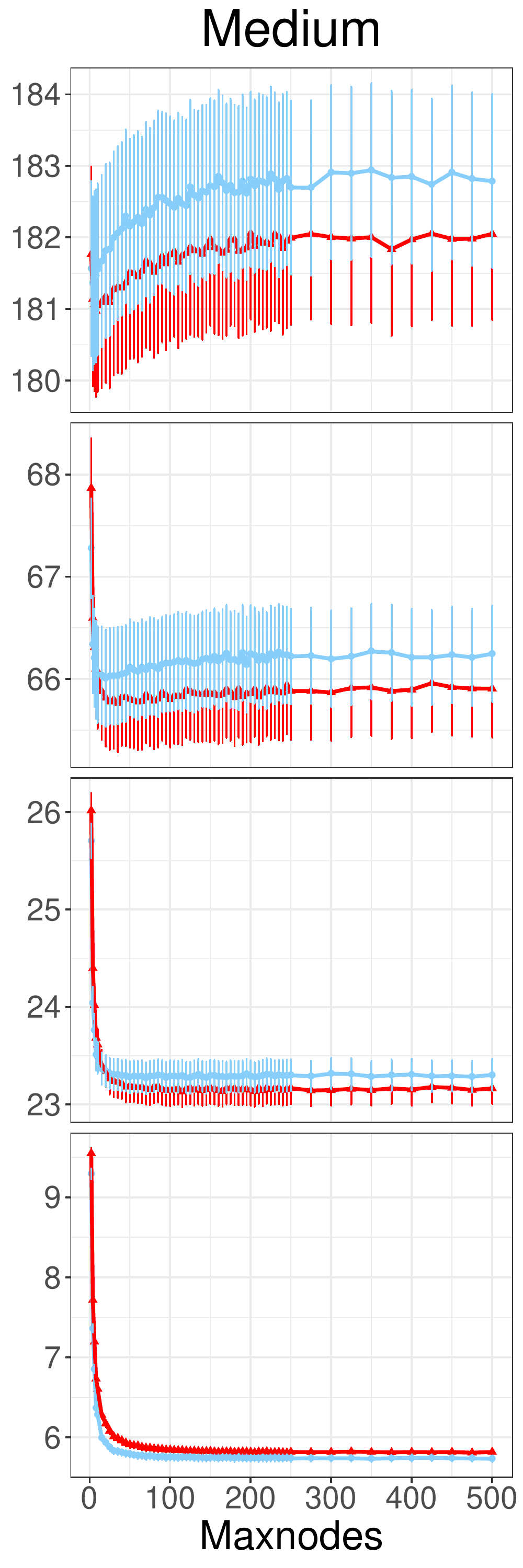}
	\includegraphics[width = 0.3\columnwidth]{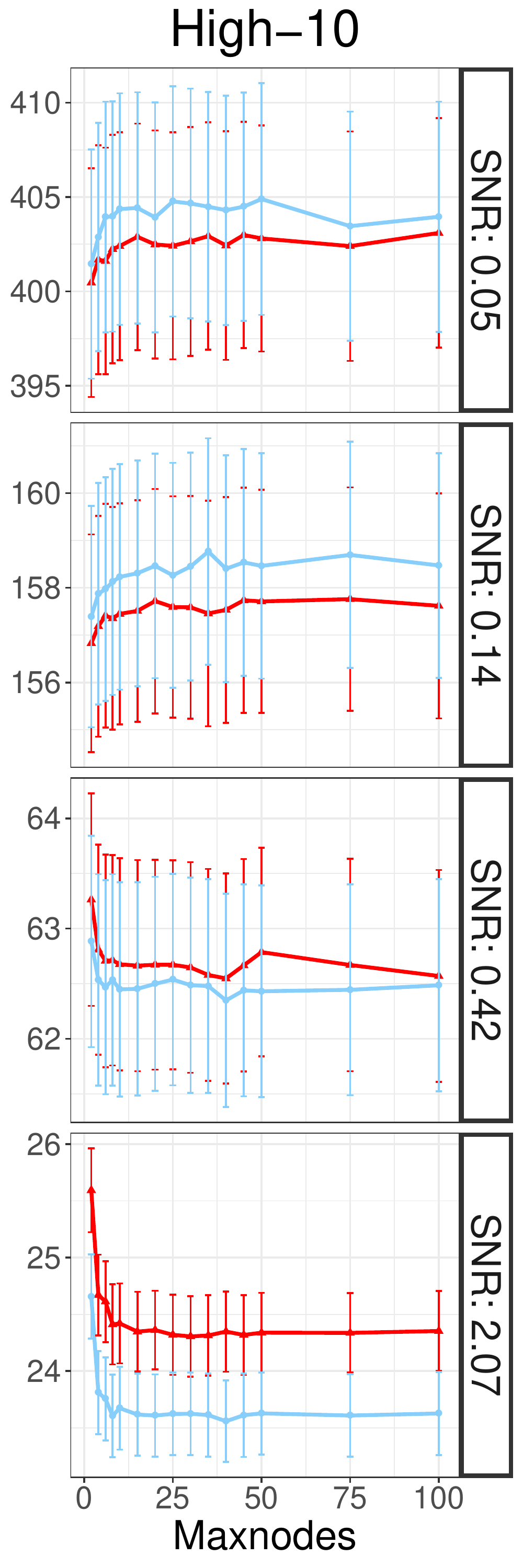}
	\caption{Performance of random forests (red) and bagging (blue) in the low (left column), medium (middle column) and high-10 settings (right column) at different SNR levels. The horizontal axis is \texttt{maxnodes}, the maximum number of nodes in each tree and vertical axis is the test MSE. Vertical bars denote one standard error.\label{fig:TreeDepth_maxnodes_SelectedSNR}}
\end{figure*}

\subsection{Simulations}
\label{sec:sims}

Here we follow closely the simulation setups utilized in recent empirical studies, including those in \cite{Hastie2020best} and \cite{mentch2020randomization}.  We assume the standard linear model relationship $Y_i = \bm{X}_i^{\top}\beta + \epsilon_i$ where the data arrive as $n$ i.i.d.\ ordered pairs $(\bm{X}_i, Y_i)$ where $\bm{X}_i \in \mathbb{R}^p$ is a vector of $p$ covariates sampled from $N_p(0, \Sigma)$ and $Y_i \in \mathbb{R}$ is the response.  Here, the $(i,j)^{th}$ entry of $\Sigma$ is of the form $\rho^{|i-j|}$ and $\rho$ is set to 0.35.  The first $s$ terms of the coefficient vector $\beta$ are set equal to 1 and the rest to 0, corresponding to the ``beta-type 2" setup in \cite{Hastie2020best}. The noise terms $\epsilon_i$ are independent of the covariates and are sampled i.i.d.\ from $N(0, \sigma^2)$ where $\sigma^2$ is chosen to create the desired signal-to-noise ratio (SNR) given by 
\[\text{SNR} = \frac{\beta^{\top}\Sigma\beta}{\sigma^2}.\]

As in the previous studies, we consider 10 SNR values ranging from 0.05 to 6, equally spaced on log scale. Under these conditions, we then consider the following four setups:

\begin{itemize}[label=\textbullet]
	\item \textbf{Low}: $n = 100$, $p=10$, $s=5$
	\vspace{1mm}
	\item \textbf{Medium}: $n = 500$, $p=100$, $s=5$
	\vspace{1mm}
	\item \textbf{High-5}: $n = 50$, $p=1000$, $s=5$
	\vspace{1mm}
	\item \textbf{High-10}: $n = 100$, $p=1000$, $s=10$
\end{itemize}

The RFs are constructed using the \texttt{randomForest} package in \texttt{R} with the \texttt{nodesize} parameter set equal to 1.  We use the \texttt{maxnodes} parameter as a proxy for tree depth and allow it to take values $\{2,4,6,8, 10, 15, \dots, n/2, n/2 + 25, n/2 + 50, \dots, n\}$.  We consider both bagging and traditional RF setups so that the \texttt{mtry} parameter is set to either $p$ or $p/3$, respectively.  Each ensemble consists of 500 trees and all remaining parameters are set to their default values. Model performance is measured as the mean squared error (MSE) evaluated on an independently generated test set the same size as the training set.  At each individual setting, the entire process is repeated 100 times and the mean and standard error of the test MSEs across these repetitions are reported. 

Results for the low, medium and high-10 setups at SNR levels of 0.05, 0.14, 0.42, and 2.07 are shown in Figure \ref{fig:TreeDepth_maxnodes_SelectedSNR} where the vertical bars represent one standard error; full results across all setups and SNR levels are given in the appendix.  In both sets of Figures, the blue curve corresponds to the results from bagging and the red to RFs.  If we look at the two higher SNR levels in the low setting (Figure \ref{fig:TreeDepth_maxnodes_SelectedSNR}, left-most column, bottom two plots), we see the pattern most would expect for both bagging and random forests -- as the trees in the ensemble grow deeper (\texttt{maxnodes} increases), the accuracy improves and eventually begins to level off.  Indeed, if this were the case across all settings, the conventional wisdom of simply growing trees in an RF to full depth would seem accurate; there would be no need to waste additional computational resources trying to optimize tree depth.  As predicted by the theory presented in the previous subsection, however, this is not the case.  At the two lowest SNR levels in the low setting, we see the opposite pattern -- shallow trees perform very well and as the depth of the trees \emph{increases}, the error increases as well before beginning to level off.  Much the same story can be seen in the medium setup (Figure \ref{fig:TreeDepth_maxnodes_SelectedSNR}, middle column) except that the switch to a preference for deeper trees appears to occur at a lower SNR.  In the high-10 setup (Figure \ref{fig:TreeDepth_maxnodes_SelectedSNR}, right-most column), a similar pattern is discernible, though as would be expected in high-dimensional settings, there appears to be a higher variance as well.  Overall across all settings, we see the same general behavior in both bagging and random forests.

\begin{figure*}
	\centering
	\includegraphics[width=0.85\textwidth]{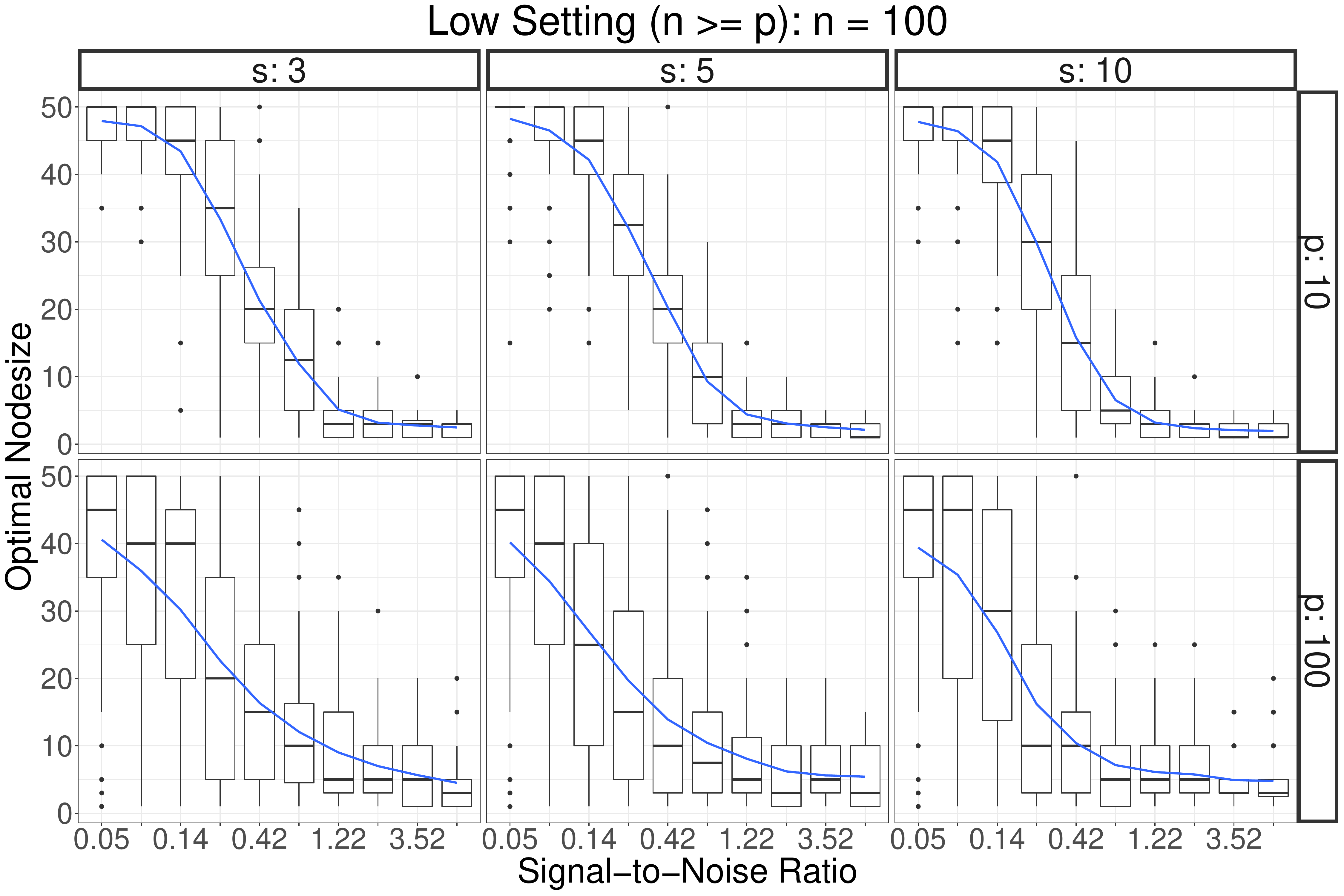}
	\caption{Boxplots of optimal \texttt{nodesize} of RFs in the Low setting ($n=100$) over 100 repetitions.}
	\label{fig:tree_depth_nodesize_low}
\end{figure*}

The results in Figure \ref{fig:TreeDepth_maxnodes_SelectedSNR} demonstrate how RF performance changes as a function of tree depth across a variety of regression setups and SNR levels.  We now formulate the question in a slightly different fashion:  given a particular data setup, what tree depth minimizes the resulting error of a RF at various SNR levels?  Here we use \texttt{nodesize} as the proxy for tree depth and allow trees to grow to the maximal possible depth subject to the \texttt{nodesize} constraint. Specifically, we consider the following three setups:  
\begin{itemize}[label=\textbullet]
	\item \textbf{Low $(n\geq p)$:  } $n=100$: $p = 10$ or 100, and $s = 3$, 5 or 10;  \texttt{nodesize} takes values 1, 3, and 10 additional values equally spaced between 5 and $n/2$.
	\vspace{1mm}
	\item \textbf{Medium $(n\geq p)$:  } $n=500$, $p = 10$ or 100, and $s = 3$, 5 or 10;  \texttt{nodesize} takes values 1, 3, and 25 additional values equally spaced between 5 and $n/2$.
	\vspace{1mm}
	\item \textbf{High $(n\leq p)$:  } $n = 50$ or 100, $p=1000$, and $s = 5$, 10 or 20;  \texttt{nodesize} takes values 1, 3, and 10 additional values equally spaced between 5 and $n/2$.
\end{itemize}

For each setting, we consider the same linear model setups and SNR values as above.  At each combination of settings, we obtain and record the optimal \texttt{nodesize}, defined as that which minimizes the MSE on an independent test set with 1000 observations.  The process is repeated 100 times and boxplots of the optimal node sizes at each SNR level in the low, medium, and high settings are shown in Figures \ref{fig:tree_depth_nodesize_low}, \ref{fig:tree_depth_nodesize_med}, and \ref{fig:tree_depth_nodesize_high}, respectively, with the latter two figures appearing in the appendix.

The results here are very much in-line with what would be expected based on the theory and experiments above.  In particular, in the low and medium settings, regardless of the number of signal covariates $s$, the same general pattern is clear.  At lower SNRs, the optimal \texttt{nodesize} is relatively large, meaning that terminal nodes in each tree often contain many observations and the trees are therefore more shallow.  As the SNR levels increase, there is an obvious transition to a preference for a smaller \texttt{nodesize} (deeper trees).  In the high setting (Figure \ref{fig:tree_depth_nodesize_high}), once again we see substantially more variance in the results, though in the larger $n$ case ($n=100$), a downward trend is still evident.  

As alluded to above, this finding is quite intuitive:  as the quality of the data improves (SNR levels rise), it makes sense that we would want to extract as much information from it as possible by growing deeper trees.  On the other hand, when the data is of poor quality (low SNR levels), this kind of overfitting can be dangerous and we might prefer to keep only the strongest, most evident patterns seen in early splits of the tree.

\section{Discussion}
\label{sec:discussion}

In this paper, we have sought to provide a principled and in-depth study into the relationship between tree depth and RF performance.  We argued that the substantial performance gains seen when transitioning from individual trees to ensembles of trees is a natural and expected by-product of averaging high-variance base models and showed that such improvements are seen even when significant training error exists and models are far from the interpolation threshold.  We then gave both theoretical and empirical evidence that optimal tree depth is ultimately a function of of the level of noise in the data.  In that sense, limiting tree depth can be seen simply as a natural form of regularization obtained by limiting model complexity.

It is important to stress, however, that bagging and random feature subsetting -- both hallmarks of classical RFs -- also perform a kind of regularization.  Thus, in practice, these ``built-in" RF regularizers may be sufficient to achieve optimal performance without needing to result to the more computationally intensive task of tuning tree depth.  This, we suspect, is why performance gains are not always seen in practice and, as a result, is likely why many textbooks continue to suggest constructing RFs with full-depth trees.  Nonetheless, this work makes clear that this is not universally the best strategy and indeed, significant performance gains \emph{can} be had by limiting tree depth.  We hope that future textbooks will provide more appropriate messaging along these lines.  \\

\section*{Acknowledgements}
This research was supported in part by the University of Pittsburgh Center for Research Computing through the resources provided. LM was partially supported by NSF DMS2015400. The MNIST data can be downloaded from \url{http://yann.lecun.com/exdb/mnist/} and loaded into \texttt{R} using the functions available at \url{https://gist.github.com/brendano/39760}. No preprocessing (other than random sampling) was done in the work above. The code necessary to reproduce the simulations and calculations in this paper is publicly available at \url{https://github.com/syzhou5/TreeDepth}.

\bibliographystyle{Chicago}
\bibliography{TreeDepth_Arxiv}

\begin{thebibliography}{}

\bibitem[\protect\citeauthoryear{Belkin, Hsu, Ma, and Mandal}{Belkin
  et~al.}{2019}]{Belkin2019}
Belkin, M., D.~Hsu, S.~Ma, and S.~Mandal (2019).
\newblock Reconciling modern machine-learning practice and the classical
  bias--variance trade-off.
\newblock {\em Proceedings of the National Academy of Sciences\/}~{\em
  116\/}(32), 15849--15854.

\bibitem[\protect\citeauthoryear{Biau and Scornet}{Biau and
  Scornet}{2016}]{Biau2016}
Biau, G. and E.~Scornet (2016).
\newblock A random forest guided tour.
\newblock {\em Test\/}~{\em 25\/}(2), 197--227.

\bibitem[\protect\citeauthoryear{Breiman}{Breiman}{1996}]{Breiman1996}
Breiman, L. (1996).
\newblock {B}agging predictors.
\newblock {\em Machine Learning\/}~{\em 24}, 123--140.

\bibitem[\protect\citeauthoryear{Breiman}{Breiman}{2001}]{Breiman2001}
Breiman, L. (2001).
\newblock {R}andom {F}orests.
\newblock {\em Machine Learning\/}~{\em 45}, 5--32.

\bibitem[\protect\citeauthoryear{Breiman, Friedman, Stone, and Olshen}{Breiman
  et~al.}{1984}]{CART}
Breiman, L., J.~Friedman, C.~J. Stone, and R.~Olshen (1984).
\newblock {\em {C}lassification and {R}egression {T}rees\/} (1st ed.).
\newblock Belmont, CA: Wadsworth.

\bibitem[\protect\citeauthoryear{B{\"u}hlmann, Yu, et~al.}{B{\"u}hlmann
  et~al.}{2002}]{buhlmann2002analyzing}
B{\"u}hlmann, P., B.~Yu, et~al. (2002).
\newblock Analyzing bagging.
\newblock {\em The Annals of Statistics\/}~{\em 30\/}(4), 927--961.

\bibitem[\protect\citeauthoryear{Duroux and Scornet}{Duroux and
  Scornet}{2018}]{Duroux2018}
Duroux, R. and E.~Scornet (2018).
\newblock Impact of subsampling and tree depth on random forests.
\newblock {\em ESAIM: PS\/}~{\em 22}, 96--128.

\bibitem[\protect\citeauthoryear{Fern{\'a}ndez-Delgado, Cernadas, Barro, and
  Amorim}{Fern{\'a}ndez-Delgado et~al.}{2014}]{Fernandez2014}
Fern{\'a}ndez-Delgado, M., E.~Cernadas, S.~Barro, and D.~Amorim (2014).
\newblock Do we need hundreds of classifiers to solve real world classification
  problems?
\newblock {\em The Journal of Machine Learning Research\/}~{\em 15\/}(1),
  3133--3181.

\bibitem[\protect\citeauthoryear{Hastie, Montanari, Rosset, and
  Tibshirani}{Hastie et~al.}{2019}]{Hastie2019}
Hastie, T., A.~Montanari, S.~Rosset, and R.~J. Tibshirani (2019).
\newblock Surprises in high-dimensional ridgeless least squares interpolation.
\newblock {\em arXiv preprint arXiv:1903.08560\/}.

\bibitem[\protect\citeauthoryear{Hastie, Tibshirani, and Friedman}{Hastie
  et~al.}{2009}]{esl}
Hastie, T., R.~Tibshirani, and J.~Friedman (2009).
\newblock {\em {T}he {E}lements of {S}tatistical {L}earning: {D}ata {M}ining,
  {I}nference, and {P}rediction\/} (2nd ed.).
\newblock New York: Springer.

\bibitem[\protect\citeauthoryear{Hastie, Tibshirani, Tibshirani, et~al.}{Hastie
  et~al.}{2020}]{Hastie2020best}
Hastie, T., R.~Tibshirani, R.~Tibshirani, et~al. (2020).
\newblock Best subset, forward stepwise or lasso? analysis and recommendations
  based on extensive comparisons.
\newblock {\em Statistical Science\/}~{\em 35\/}(4), 579--592.

\bibitem[\protect\citeauthoryear{Hoeffding}{Hoeffding}{1948}]{HoeffdingUstat}
Hoeffding, W. (1948).
\newblock A {C}lass of {S}tatistics with {A}symptotically {N}ormal
  {D}istribution.
\newblock {\em The Annals of Mathematical Statistics\/}~{\em 19\/}(3),
  293--325.

\bibitem[\protect\citeauthoryear{Izenman}{Izenman}{2008}]{Izenman2008modern}
Izenman, A.~J. (2008).
\newblock {\em Modern multivariate statistical techniques}, Volume~10.
\newblock Springer.

\bibitem[\protect\citeauthoryear{James, Witten, Hastie, and Tibshirani}{James
  et~al.}{2013}]{James2013introduction}
James, G., D.~Witten, T.~Hastie, and R.~Tibshirani (2013).
\newblock {\em An introduction to statistical learning}, Volume 112.
\newblock Springer.

\bibitem[\protect\citeauthoryear{LeJeune, Javadi, and Baraniuk}{LeJeune
  et~al.}{2020}]{Lejeune2020implicit}
LeJeune, D., H.~Javadi, and R.~Baraniuk (2020).
\newblock The implicit regularization of ordinary least squares ensembles.
\newblock In {\em International Conference on Artificial Intelligence and
  Statistics}, pp.\  3525--3535. PMLR.

\bibitem[\protect\citeauthoryear{Lin and Jeon}{Lin and Jeon}{2006}]{Lin2006}
Lin, Y. and Y.~Jeon (2006).
\newblock Random forests and adaptive nearest neighbors.
\newblock {\em Journal of the American Statistical Association\/}~{\em
  101\/}(474), 578--590.

\bibitem[\protect\citeauthoryear{Mei and Montanari}{Mei and
  Montanari}{2019}]{mei2019generalization}
Mei, S. and A.~Montanari (2019).
\newblock The generalization error of random features regression: Precise
  asymptotics and double descent curve.
\newblock {\em arXiv preprint arXiv:1908.05355\/}.

\bibitem[\protect\citeauthoryear{Mentch and Hooker}{Mentch and
  Hooker}{2016}]{Mentch2016}
Mentch, L. and G.~Hooker (2016).
\newblock Quantifying uncertainty in random forests via confidence intervals
  and hypothesis tests.
\newblock {\em The Journal of Machine Learning Research\/}~{\em 17\/}(1),
  841--881.

\bibitem[\protect\citeauthoryear{Mentch and Hooker}{Mentch and
  Hooker}{2017}]{Mentch2017}
Mentch, L. and G.~Hooker (2017).
\newblock Formal hypothesis tests for additive structure in random forests.
\newblock {\em Journal of Computational and Graphical Statistics\/}~{\em
  26\/}(3), 589--597.

\bibitem[\protect\citeauthoryear{Mentch and Zhou}{Mentch and
  Zhou}{2020}]{mentch2020randomization}
Mentch, L. and S.~Zhou (2020).
\newblock Randomization as regularization: A degrees of freedom explanation for
  random forest success.
\newblock {\em Journal of Machine Learning Research\/}~{\em 21\/}(171), 1--36.

\bibitem[\protect\citeauthoryear{Probst, Wright, and Boulesteix}{Probst
  et~al.}{2019}]{Probst2019}
Probst, P., M.~N. Wright, and A.-L. Boulesteix (2019).
\newblock Hyperparameters and tuning strategies for random forest.
\newblock {\em Wiley Interdisciplinary Reviews: Data Mining and Knowledge
  Discovery\/}~{\em 9\/}(3), e1301.

\bibitem[\protect\citeauthoryear{Scornet, Biau, Vert, et~al.}{Scornet
  et~al.}{2015}]{Scornet2015}
Scornet, E., G.~Biau, J.-P. Vert, et~al. (2015).
\newblock Consistency of random forests.
\newblock {\em The Annals of Statistics\/}~{\em 43\/}(4), 1716--1741.

\bibitem[\protect\citeauthoryear{Segal}{Segal}{2004}]{segal2004machine}
Segal, M.~R. (2004).
\newblock Machine learning benchmarks and random forest regression.

\bibitem[\protect\citeauthoryear{Wager and Athey}{Wager and
  Athey}{2018}]{Wager2018}
Wager, S. and S.~Athey (2018).
\newblock Estimation and inference of heterogeneous treatment effects using
  random forests.
\newblock {\em Journal of the American Statistical Association\/}~{\em
  113\/}(523), 1228--1242.

\bibitem[\protect\citeauthoryear{Wyner, Olson, Bleich, and Mease}{Wyner
  et~al.}{2017}]{Wyner2017}
Wyner, A.~J., M.~Olson, J.~Bleich, and D.~Mease (2017).
\newblock Explaining the success of adaboost and random forests as
  interpolating classifiers.
\newblock {\em The Journal of Machine Learning Research\/}~{\em 18\/}(1),
  1558--1590.

\end{thebibliography}

\appendix
\newpage

\section*{Appendix: Additional Figures}
\label{sec:appendix_figures}

\begin{figure}[!ht]
	\centering
	\includegraphics[width=0.85\textwidth]{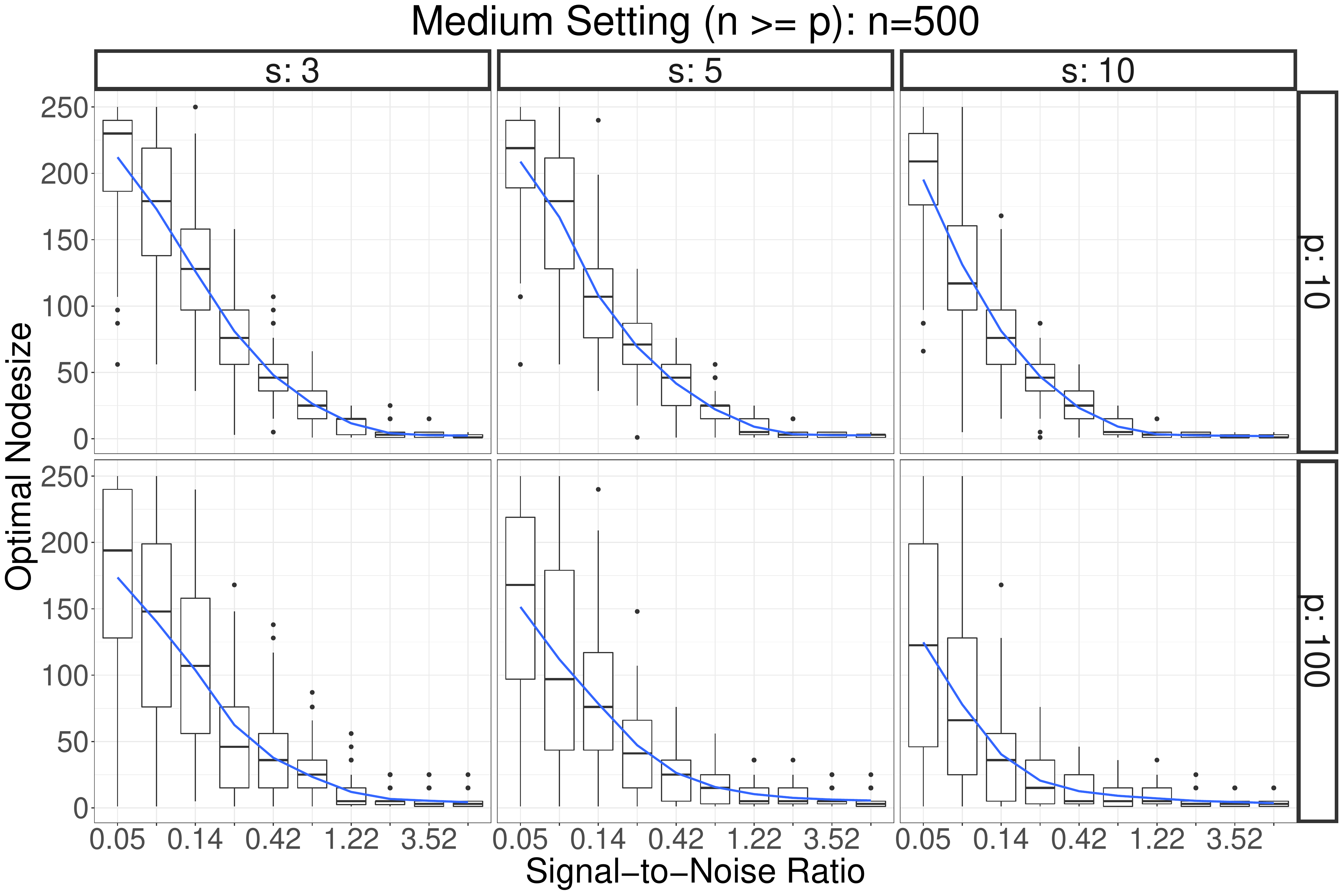}
\caption{Boxplots of optimal \texttt{nodesize} of RFs in the Meidum setting ($n=500$) over 100 repetitions.}
	\label{fig:tree_depth_nodesize_med}
\end{figure}

\begin{figure}[!ht]
	\centering
	\includegraphics[width=0.85\textwidth]{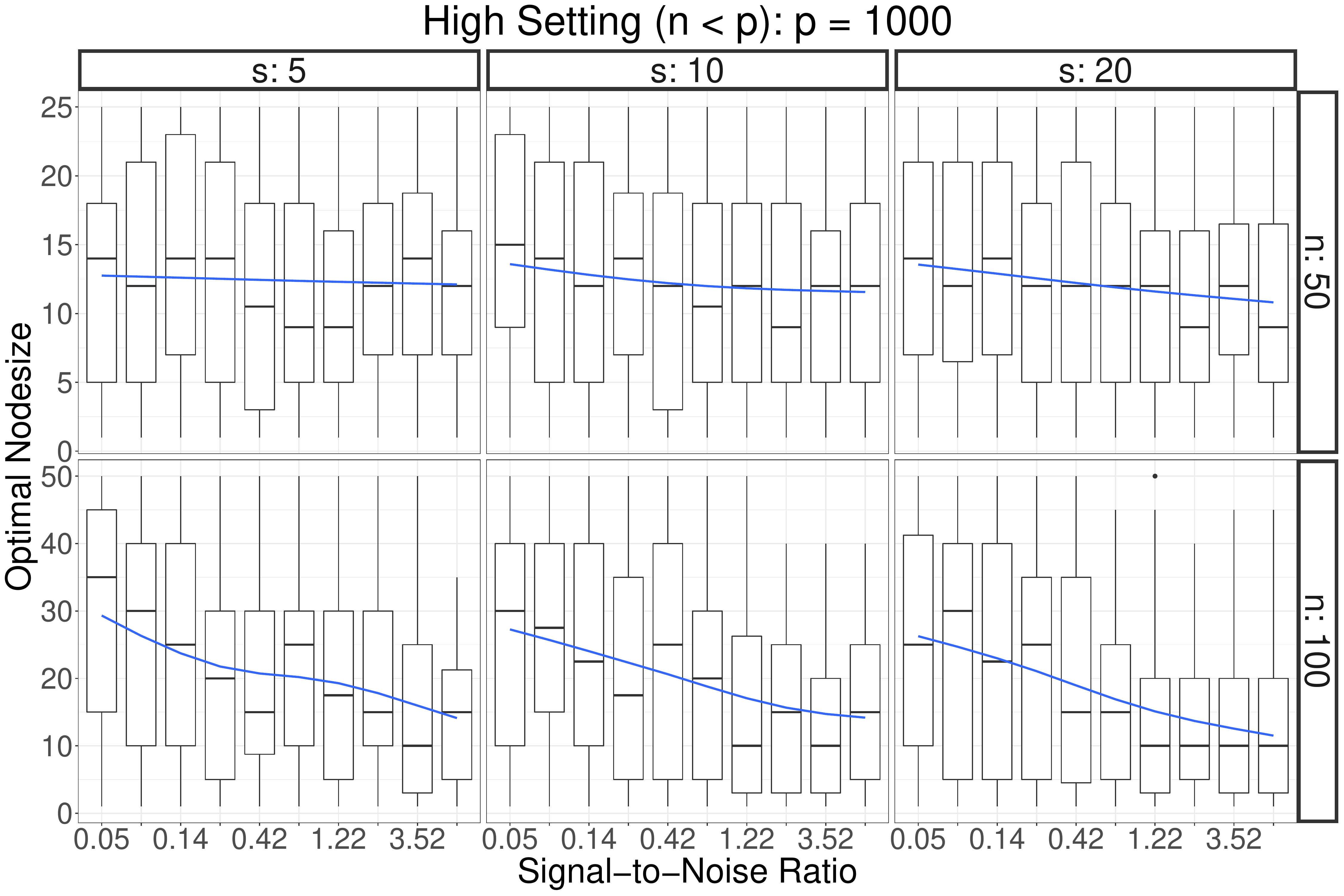}
	\caption{Boxplots of optimal \texttt{nodesize} of RFs in the High setting ($p=1000$) over 100 repetitions.}
	\label{fig:tree_depth_nodesize_high}
\end{figure}

\begin{figure}[!ht]
	\centering
	\includegraphics[width = 0.95\textwidth]{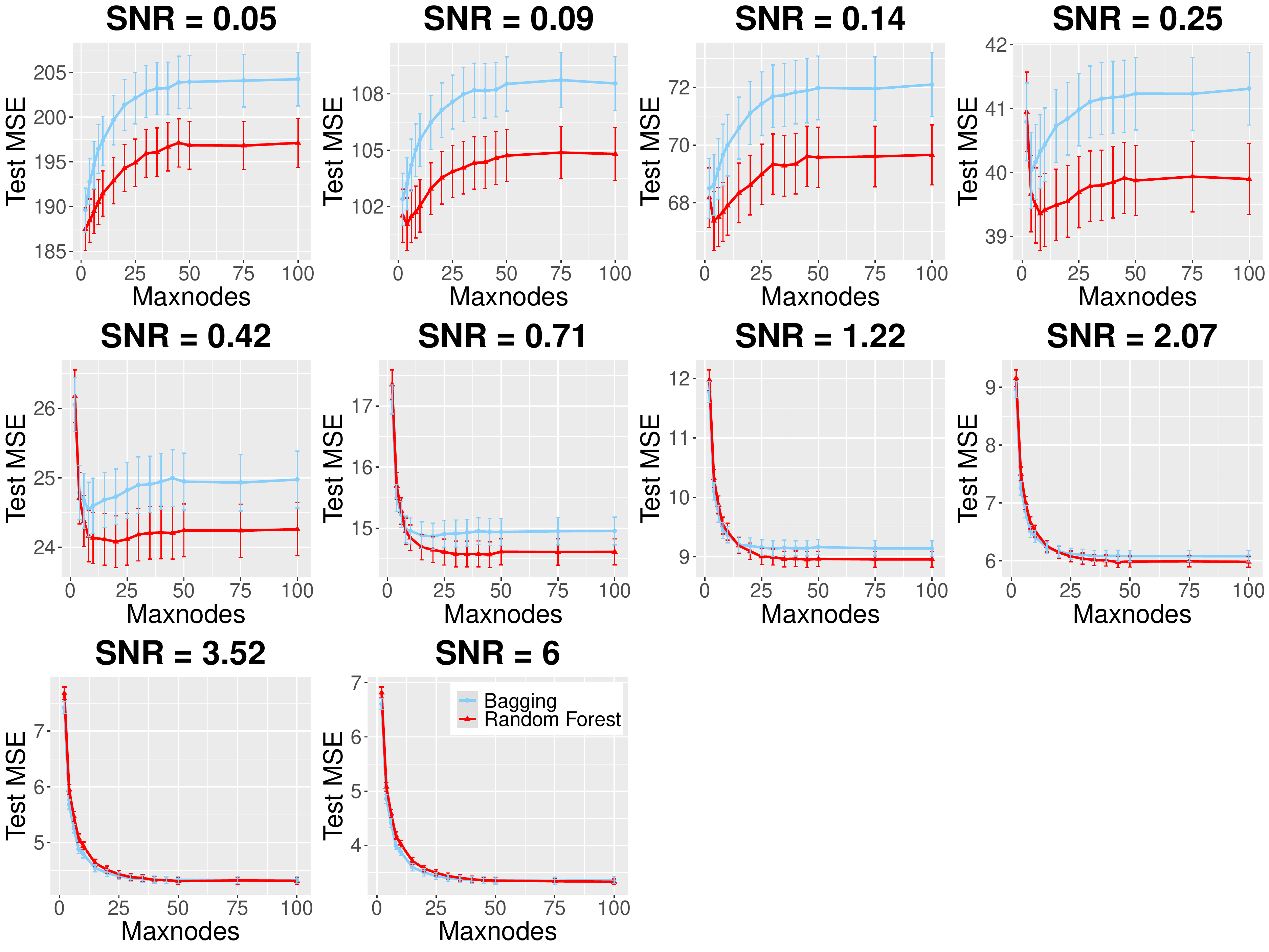}
	\caption{Performance of RFs with \texttt{mtry} equal to $p/3$ (the default value) and $p$ (bagging) in the \textbf{low} setting. Vertical bars denote one standard error.}
	\label{fig:TreeDepth_maxnodes_low}
\end{figure}

\begin{figure}[!ht]
	\centering
	\includegraphics[width = 0.95\textwidth]{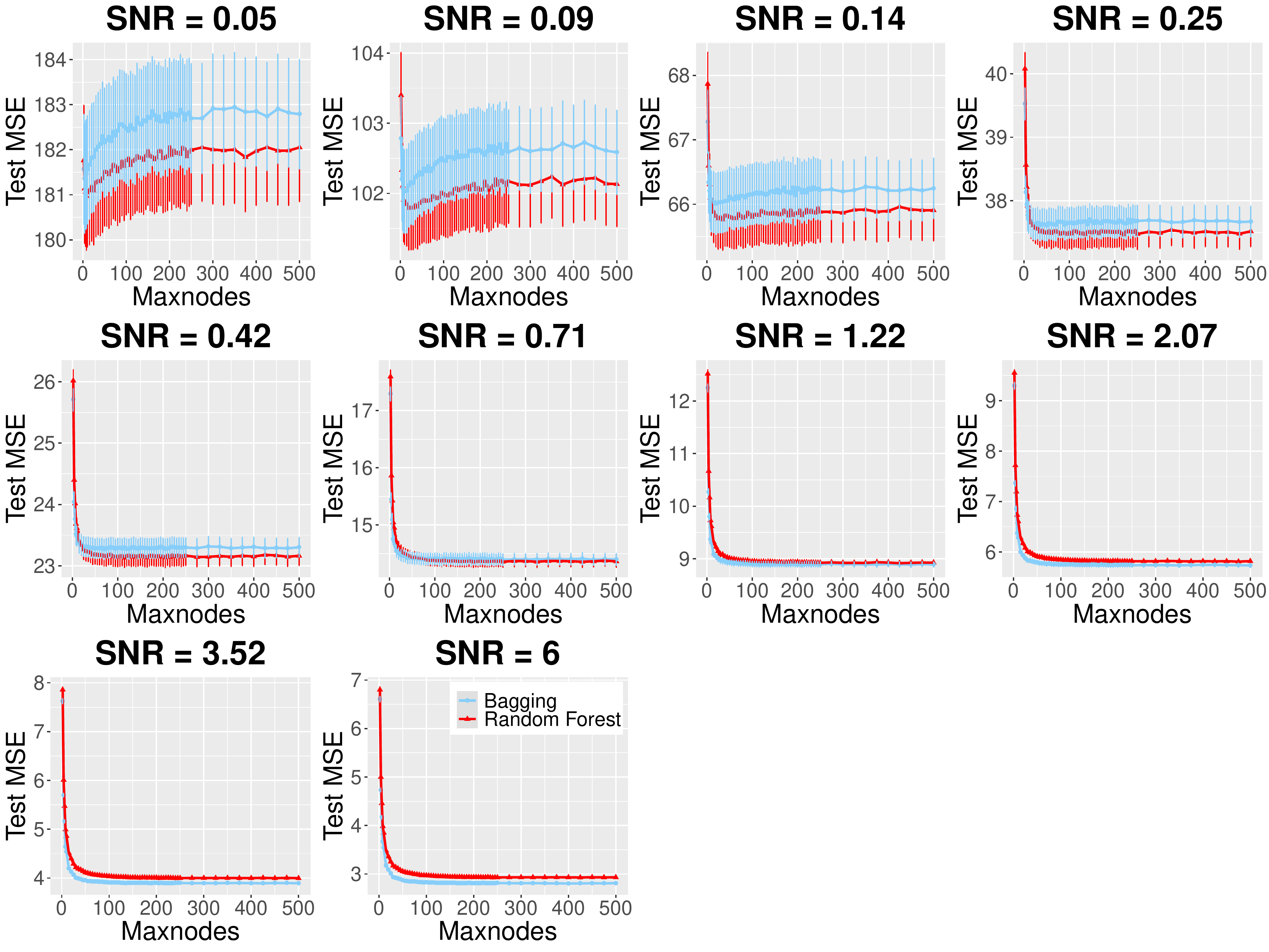}
\caption{Performance of RFs with \texttt{mtry} equal to $p/3$ (the default value) and $p$ (bagging) in the \textbf{medium} setting. Vertical bars denote one standard error.}
	\label{fig:TreeDepth_maxnodes_med}
\end{figure}

\begin{figure}[!ht]
	\centering
	\includegraphics[width = 0.95\textwidth]{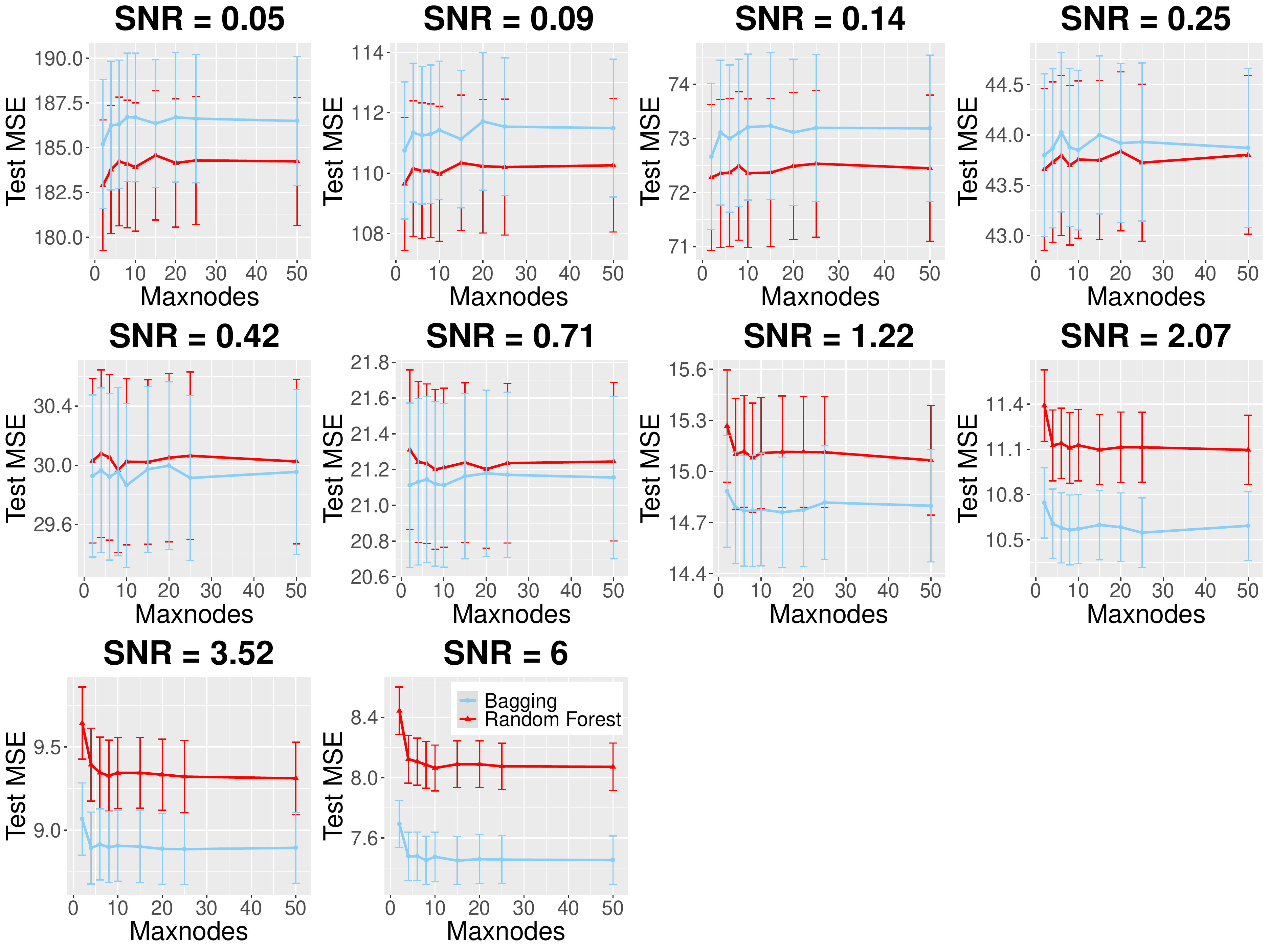}
\caption{Performance of RFs with \texttt{mtry} equal to $p/3$ (the default value) and $p$ (bagging) in the \textbf{high-5} setting. Vertical bars denote one standard error.}
	\label{fig:TreeDepth_maxnodes_high5}
\end{figure}

\begin{figure}[!ht]
	\centering
	\includegraphics[width = 0.95\textwidth]{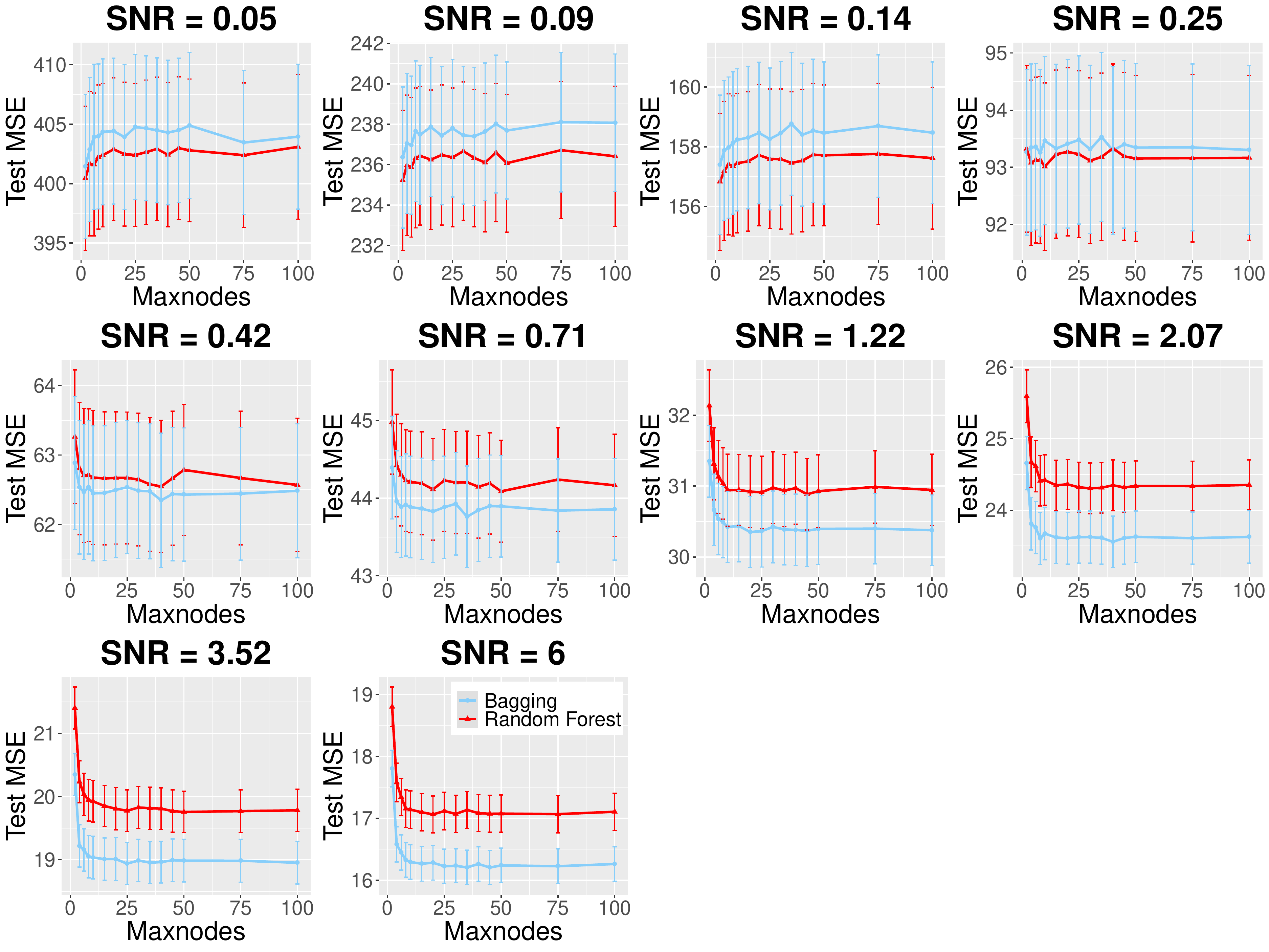}
\caption{Performance of RFs with \texttt{mtry} equal to $p/3$ (the default value) and $p$ (bagging) in the \textbf{high-10} setting. Vertical bars denote one standard error.}
	\label{fig:TreeDepth_maxnodes_high10}
\end{figure}

Figures \ref{fig:tree_depth_nodesize_med} and \ref{fig:tree_depth_nodesize_high} show the relationship between optimal \texttt{nodesize} and SNR in the medium and high settings, respectively, described in Section \ref{sec:sims}.  Figures \ref{fig:TreeDepth_maxnodes_low}--\ref{fig:TreeDepth_maxnodes_high10} show RF performance vs \texttt{maxnodes} across the full range of all 10 SNRs under investigation in the low, medium, high-5, and high-10 settings also described in Section \ref{sec:sims}.  Note that Figure \ref{fig:TreeDepth_maxnodes_SelectedSNR} was designed to show a representative subset of these and was included for the purpose of preserving space in the main text.

\end{document}